\documentclass[letterpaper, 10 pt, conference]{ieeeconf}  

\IEEEoverridecommandlockouts                              

\usepackage[left=1.88cm,right=1.88cm,top=1.88cm,bottom=1.88cm]{geometry}
\usepackage{amsmath,amssymb,amsfonts}
\usepackage{cite}
\usepackage{graphicx}
\usepackage{textcomp}
\usepackage{xcolor}
\usepackage{soul}
\usepackage{bbm}
\let\labelindent\relax
\usepackage{enumitem}
\usepackage{tabularx}

\usepackage{times}
\usepackage{multicol}

\usepackage{graphicx}
\usepackage{bm}
\usepackage[nolist]{acronym}

\usepackage[linesnumbered,ruled]{algorithm2e}

\usepackage{mathtools}
\usepackage{array}
\usepackage{dblfloatfix}
\usepackage{diagbox}
\usepackage{etoolbox}\AtBeginEnvironment{algorithmic}{\small}

\usepackage[normalem]{ulem}
\usepackage{comment}
\newif\ifshowcomments
\showcommentsfalse 

\ifshowcomments
    \newcommand{\bae}[1]{\hl{[SB: #1]}\protect\color{black}} 
    \newcommand{\di}[1]{\hl{[DI: #1]}\protect\color{black}} 
    \newcommand{\ft}[1]{\hl{[FT: #1]}\protect\color{black}} 
    \newcommand{\jd}[1]{\hl{[JD: #1]}\protect\color{black}}
    \newcommand{\lz}[1]{\hl{[LZ: #1]}\protect\color{black}}
    
    \newcommand{\amc}[1]{\hl{[AC: #1]}\protect\color{blue}}
    \newcommand{\removed}[1]{{\color{red}\sout{#1}}}  
    \specialcomment{removedblock}{\begingroup\color{red}}{\endgroup}
\else
    \newcommand{\bae}[1]{}
    \newcommand{\di}[1]{}
    \newcommand{\ft}[1]{}
    \newcommand{\jd}[1]{}
    \newcommand{\lz}[1]{}
    \newcommand{\amc}[1]{}
    \newcommand{\removed}[1]{}    
    \excludecomment{removedblock}
\fi

\newif\ifshowrebuttal
\showrebuttalfalse  
\ifshowrebuttal
    \newcommand{\rnew}[1]{{\color{blue}#1}}              
    \newcommand{\rremoved}[1]{{\color{red}\sout{#1}}}    
    \newcommand{\rreplaced}[2]{{\color{red}\sout{#1}}{\color{blue}#2}} 
    \specialcomment{rnewblock}{\begingroup\color{blue}}{\endgroup}
    \specialcomment{rremovedblock}{\begingroup\color{red}}{\endgroup}
\else
    \newcommand{\rnew}[1]{#1}
    \newcommand{\rremoved}[1]{}
    \newcommand{\rreplaced}[2]{#2}
    \excludecomment{rnewblock}
    \excludecomment{rremovedblock}
\fi

\usepackage{tikz}
\usetikzlibrary{shapes.geometric, arrows.meta, positioning, fit, calc, backgrounds}
\usepackage{booktabs}
\usepackage{multirow}
\usepackage{makecell}
\usepackage{graphicx}        
\usepackage{subcaption}    
\usepackage{subcaption}
\usepackage[table]{xcolor}
\usepackage{enumitem} 
\usepackage[export]{adjustbox}
\usepackage{graphicx}
\usepackage{subcaption}
\usepackage{tikz}
\usepackage{tcolorbox}
\usepackage{adjustbox}
\usetikzlibrary{positioning, shapes.geometric, arrows.meta, backgrounds, fit}
\usepackage{graphicx}
\usepackage{tikz}
\usepackage{amsmath}
\usepackage{amsfonts}

\usetikzlibrary{shapes.geometric, arrows.meta, positioning, backgrounds, quotes}

\usepackage{lipsum}
\usepackage[colorinlistoftodos]{todonotes}

\usepackage{amsthm}

\usepackage{bbm} 
\usepackage[colorlinks = True, linkcolor = blue, citecolor = blue]{hyperref}

\title{\LARGE \bf
What’s Hidden Matters: Identifying Planning-Critical Occluded Agents using Vision-Language Models
 }

\newif\ifanonymous
\anonymousfalse
\ifanonymous
    \author{Anonymous Authors}
\else
    \author{Amirhosein Chahe$^{1,2}$ \quad  Tyler Naes$^1$ \quad  Jovin D'sa$^1$ \quad Faizan M. Tariq$^1$  \quad Sangjae Bae$^1$ \\ Lifeng Zhou$^2$ \quad David Isele$^1$
    \thanks{$^{1}$Honda Research Institute (HRI), San Jose, CA 95134, USA.}
    \thanks{$^{2}$Drexel University, Philadelphia, PA 19104, USA.}
    \thanks{All work was done while A. Chahe was employed by HRI.
    Contact: \href{mailto:ac4462@drexel.edu}{\texttt{ac4462@drexel.edu}},
    \href{mailto:sbae@honda-ri.com}{\texttt{sbae@honda-ri.com}},
    \href{mailto:disele@honda-ri.com}{\texttt{disele@honda-ri.com}}.}
    \thanks{\copyright 2026 IEEE. Personal use of this material is permitted.
    Permission from IEEE must be obtained for all other uses, in any current
    or future media, including reprinting/republishing this material for
    advertising or promotional purposes, creating new collective works, for
    resale or redistribution to servers or lists, or reuse of any copyrighted
    component of this work in other works.}
    }
\fi

\IEEEaftertitletext{\vspace{-1.0\baselineskip}}

\begin{document}

\maketitle


\begin{abstract}

Autonomous vehicles must safely navigate complex environments where planning-critical agents may be hidden from view. Current approaches often treat all occlusions with uniform conservatism, yielding needlessly defensive driving, or they infer hidden spaces without estimating the impact on the planner. This work bridges the critical gap between perception and planning by enabling Vision-Language Models (VLMs) to identify and reason about the specific hidden agents that are most critical to the ego-vehicle's trajectory. 
We introduce a novel framework that uses Planning KL-divergence (PKL), an information-theoretic metric, to systematically identify and rank occluded agents based on their impact on the ego vehicle's plan. Using this planning-aware ranking, we employ an expert VLM (GPT-5) to generate rich, structured annotations that capture the visual evidence and reasoning required for this task. 
We apply this framework to the nuScenes dataset to create a new benchmark focused on high-impact scenarios.
We conduct comprehensive experiments on a wide range of general-purpose and domain-adapted VLMs, demonstrating that fine-tuning on our PKL-guided data yields dramatic performance improvements across all models. Notably, our results show that smaller, fine-tuned models significantly outperform their much larger zero-shot counterparts, and that our PKL-guided data selection strategy improves performance by approximately 30\% over random sampling. Our work presents the first systematic approach for training VLMs to focus on planning-critical occlusions, enabling more semantically grounded and efficient risk assessment in autonomous driving.

\end{abstract}

\definecolor{boxblue}{RGB}{23,114,219}
\definecolor{textgreen}{RGB}{52,143,89}
\definecolor{textred}{RGB}{204,51,51}
\definecolor{pklorange}{RGB}{227,131,33}
\definecolor{gptbluebg}{RGB}{229, 239, 252} 
\definecolor{gptblueborder}{RGB}{177, 206, 242} 
\definecolor{myblue}{RGB}{33, 98, 131}
\definecolor{mygreen}{RGB}{117, 166, 64}
\definecolor{mypurple}{RGB}{104, 34, 139}

\begin{figure*}[!t]
    \centering
    \vspace{-0.2em}
    \begin{tikzpicture}[
        node distance=5mm and 5mm, 
        img_block/.style={inner sep=0pt},
        arrow/.style={-Latex, thick, draw=black!80},
        mybluearrow/.style={-Latex, thick, draw=myblue},
        mypurplearrow/.style={-Latex, thick, draw=mypurple},
        text_output_block/.style={
            rectangle,
            rounded corners,
            fill=gptbluebg,
            draw=gptblueborder,
            text width=7.00cm, 
            font=\scriptsize,
            inner sep=8pt
        }
    ]

    \node[img_block, draw=black, line width=0.5pt, rounded corners=3pt, inner sep=2pt] (nuscenes) {
    \begin{tabular}{@{}c@{}c@{}} 
        \begin{tabular}[t]{@{}c@{}} 
            \includegraphics[width=0.34\textwidth, trim={10pt 10pt 10pt 10pt}, clip]{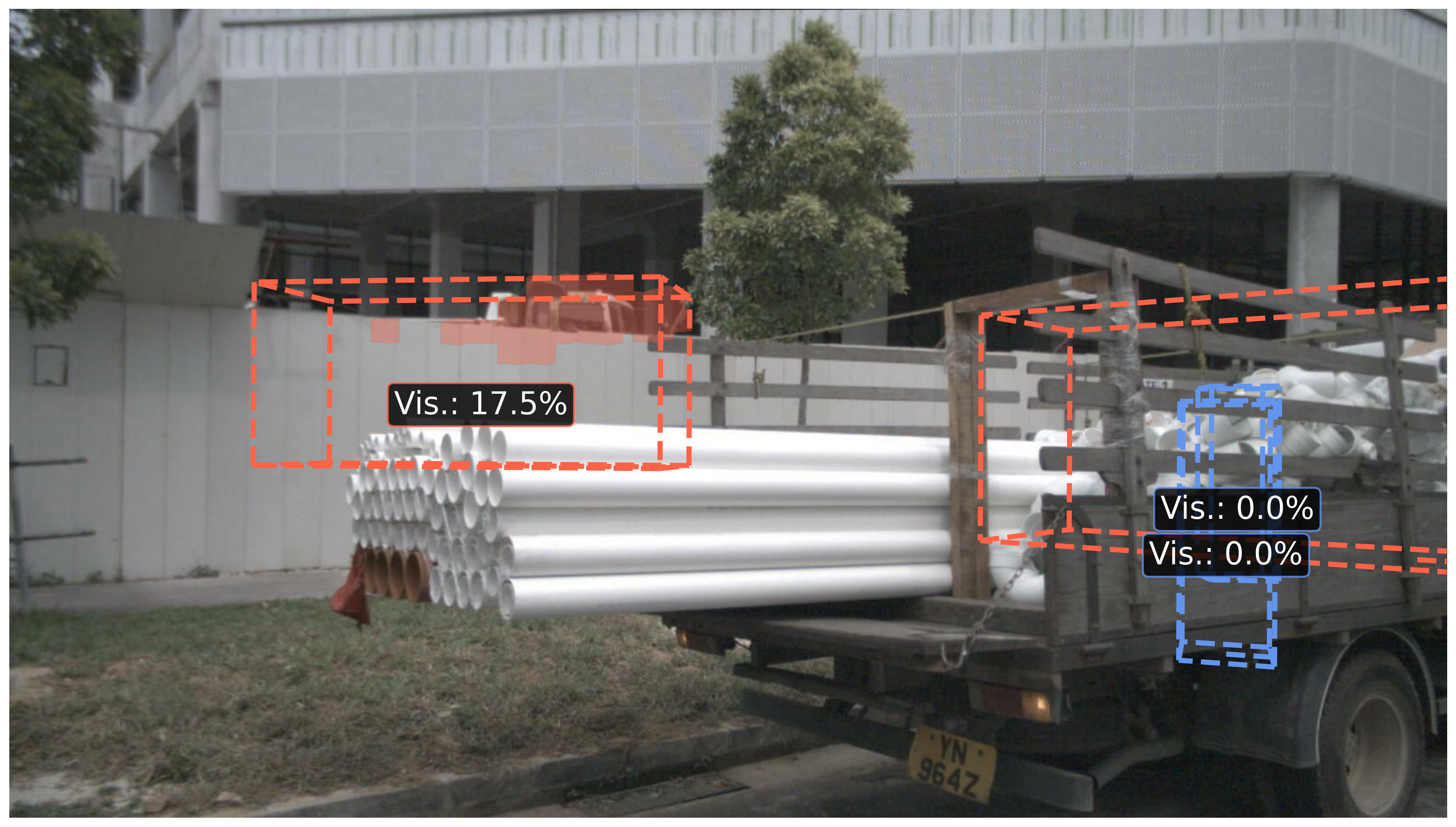} \\[-5pt]
            \footnotesize Front Left Camera 
        \end{tabular}
        &
        \begin{tabular}[t]{@{}c@{}}
            \includegraphics[width=0.34\textwidth, trim={10pt 10pt 10pt 10pt}, clip]{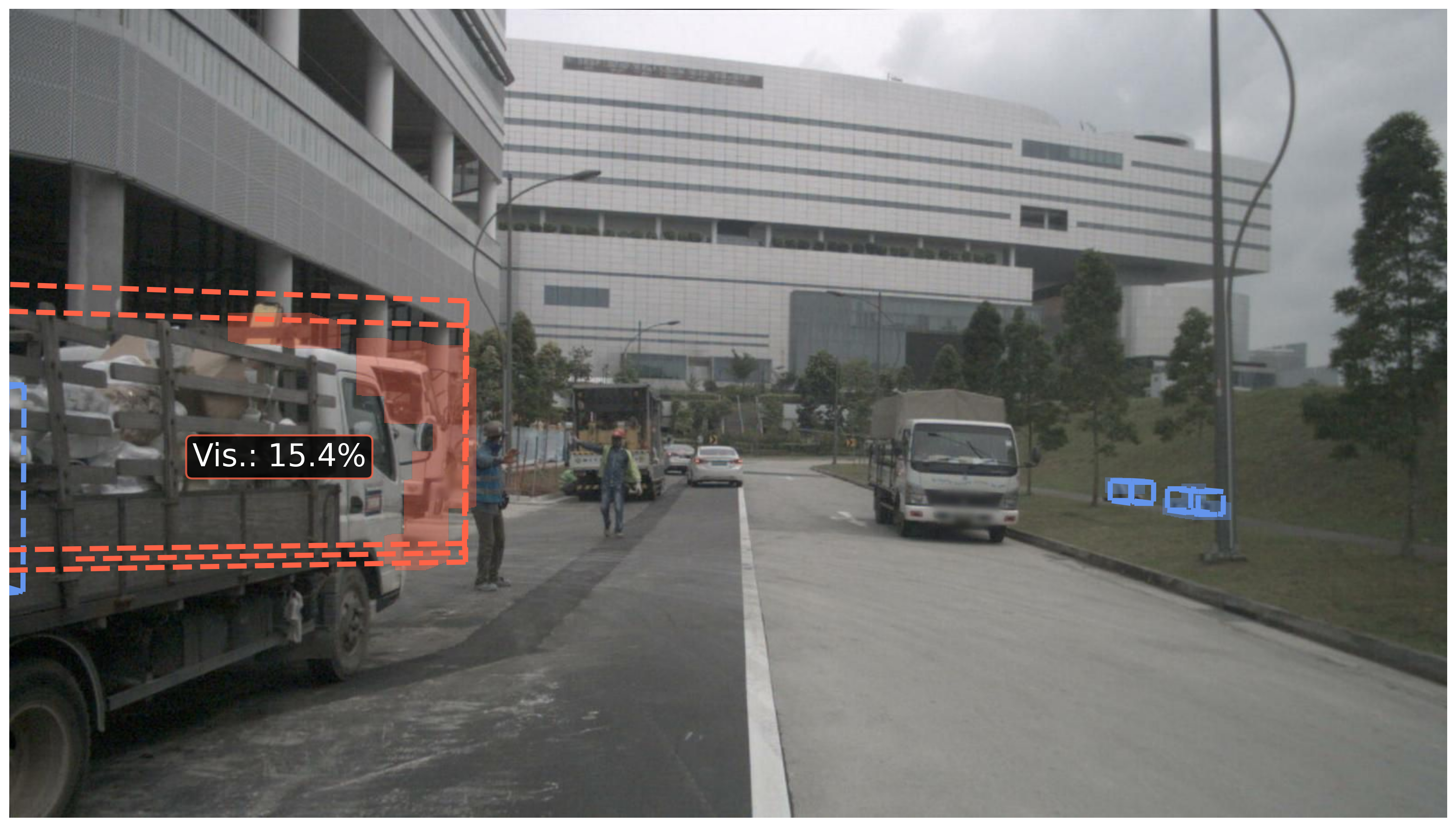} \\[-5pt]
            \footnotesize Front Camera 
        \end{tabular}
    \end{tabular}
   };
    \node[above=-1.2mm of nuscenes,xshift=-4.50cm,fill=black!80,rounded corners=3pt,inner sep=2pt, text=white, font=\small] {nuScenes + Occ3D};

    \node[img_block, right=of nuscenes, draw=black, line width=0.5pt, rounded corners=3pt, inner sep=2pt] (pkl) {
        \includegraphics[width=0.25\textwidth]{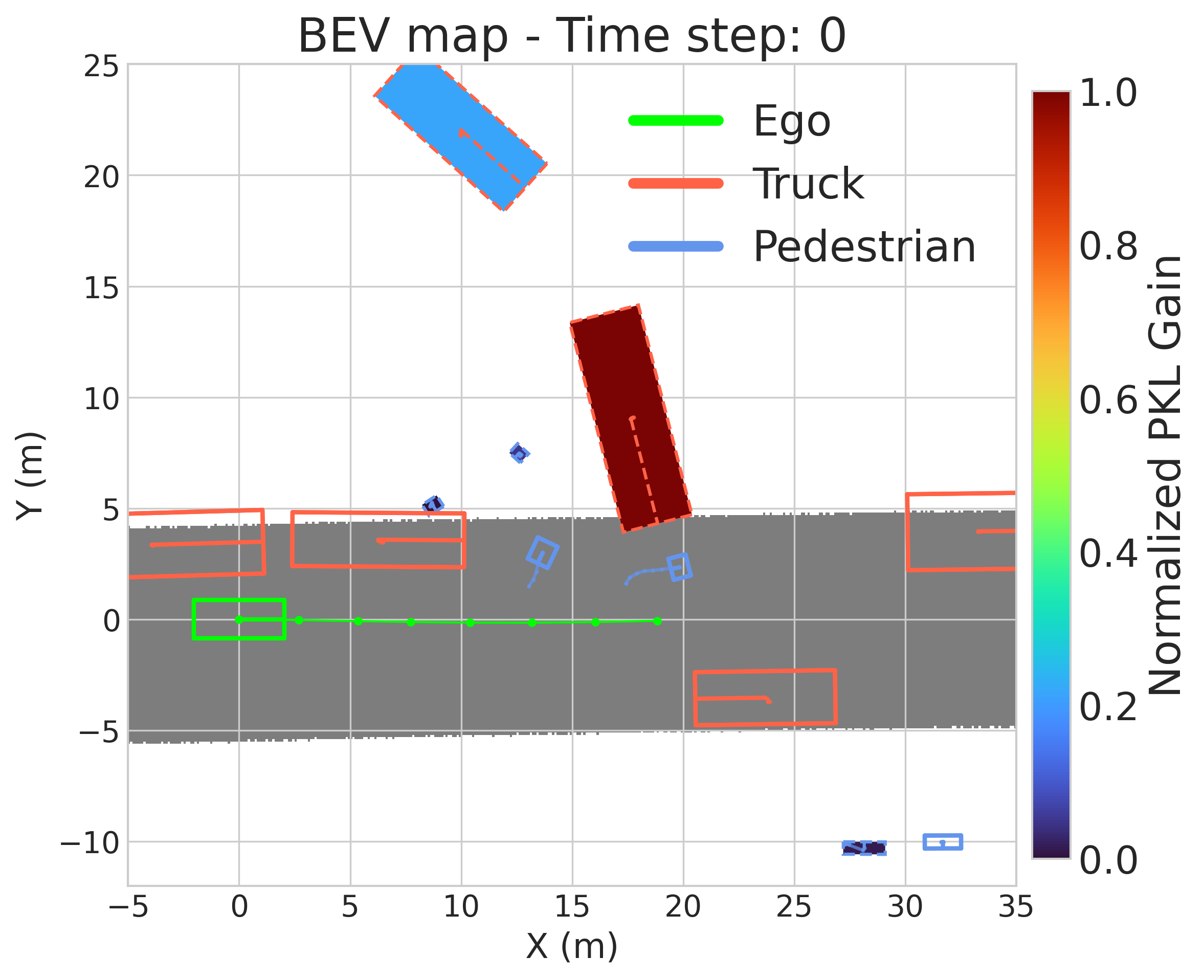}
    };
    \node[above=-1.0mm of pkl, xshift=-0.70cm, fill=black!80,rounded corners=3pt,inner sep=2pt, font=\small\color{pklorange}] {PKL-based ranking};

    \node[img_block, below=of pkl, xshift=-1.50cm, draw=black, line width=0.5pt, rounded corners=3pt, inner sep=1pt] (pano) {
        \begin{tabular}{@{}c@{}}
            \includegraphics[width=0.45\textwidth]{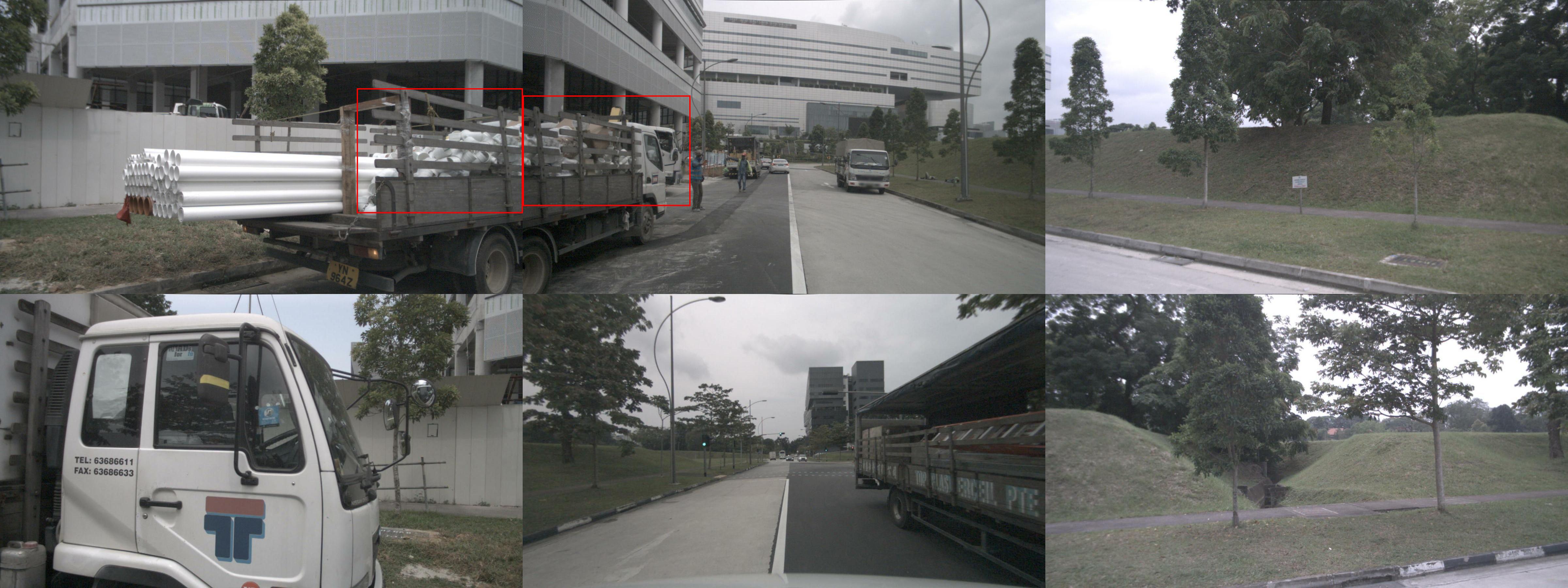}\\
            \includegraphics[width=0.025\textwidth]{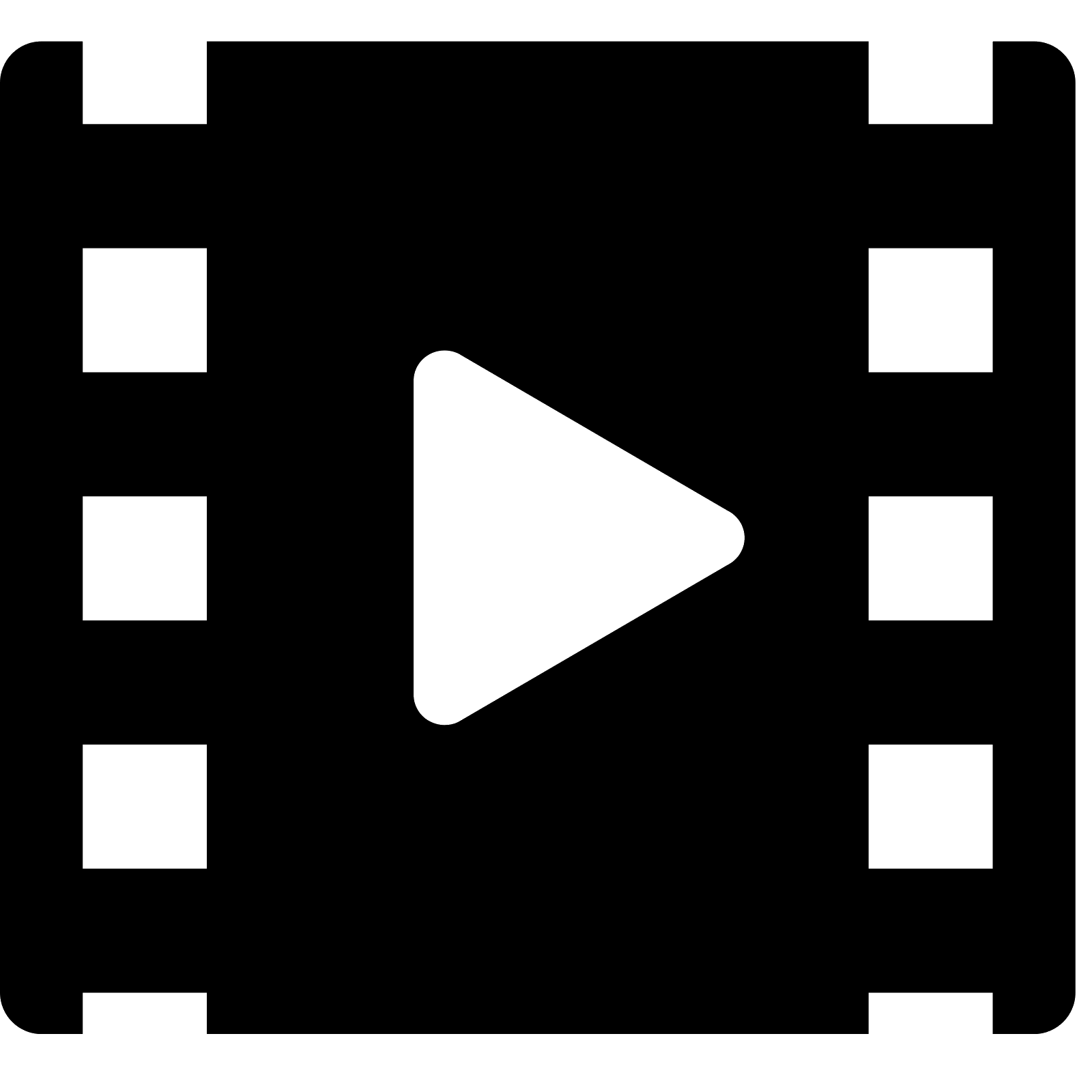}\\
            \includegraphics[width=0.45\textwidth]{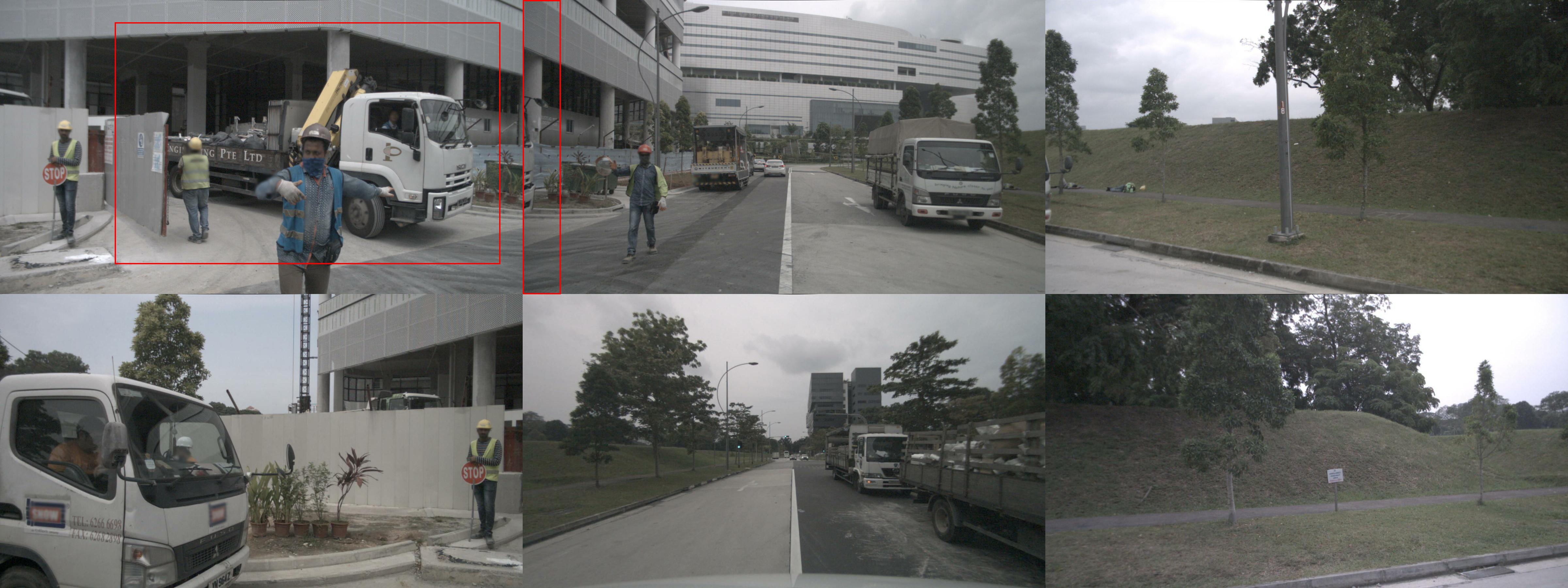}
        \end{tabular}
    };

\node[font=\footnotesize\bfseries, fill=white, inner sep=1pt, opacity=0.9] at ([xshift=-15pt, yshift=-15pt]pano.north east) {(t=0)};
\node[font=\footnotesize\bfseries, fill=white, inner sep=1pt, opacity=0.9] at ([xshift=-15pt, yshift=15pt]pano.south east) {(t=2)};

\node[fill=myblue, text=white, align=left, below=1.95cm of pkl, xshift=-10.0cm, inner sep=2pt, anchor=south, font=\tiny, rounded corners, text width=8cm] (list) {

\textbf{Prompt:} You are looking for a \texttt{\{agent\_class\}} (= truck) that may \textbf{NOT} be clearly visible in one or more timesteps and its bounding boxes are drawn in red color.\\
YOUR TASK:
\begin{itemize}
    \item Find DIRECT or INDIRECT EVIDENCE of this hidden \texttt{\{agent\_class\}} (= truck) ...
    \item Identify WHAT IS BLOCKING this agent ...
    \item Set priority/sector/reason based on ego-path visibility impact;
\end{itemize}
... return only schema-valid JSON (exact enums) ...
};

    \node[img_block, left=of pano, yshift=-1.0cm, xshift=-0.2cm] (gpt) {
        \includegraphics[width=0.03\textwidth, trim={40cm 0 40cm 0}]{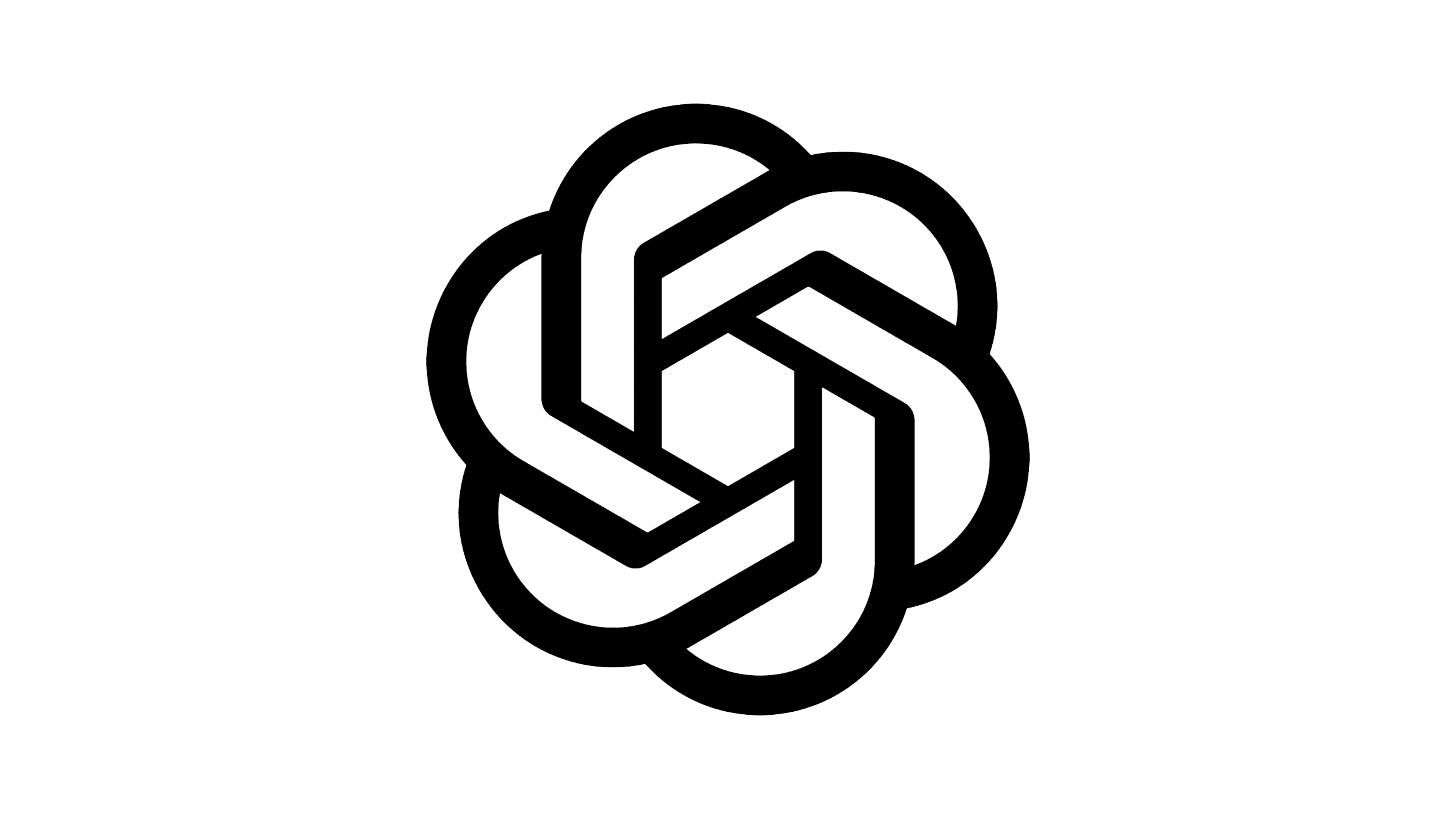}
    };
    \node[below=-1mmof gpt, font=\small] {GPT-5};

    \node[text_output_block, left=of gpt, xshift=-0.2cm] (output_text) {
        \textbf{Visual Evidence (timestep 0):} \\[1mm]
        Camera: FRONT\\
        Cue: \textcolor{textgreen}{Partial visibility} - 
        \textcolor{textgreen}{White cab edges and mirror peek between the side slats} of a \textcolor{boxblue}{nearer flatbed along the curb}, hinting at another vehicle tucked behind near the site entrance. \\
        Agent Class: \textcolor{textgreen}{Truck} \\
        Action: \textbf{Preparing to pull out} from the gate \\
        Occluder Class: \textcolor{boxblue}{Truck}\\
        Occlusion analysis: \textcolor{boxblue}{Occluded by agent} - The \textcolor{boxblue}{long flatbed parked on the curb} blocks most of the cab.\\
        ...
        \\[1mm]
        
        \textbf{Priority:} \\
        Level: \textcolor{textred}{High} \\
        Sector: Ahead-Left \\
        Reason: Truck emerging from a construction gate on the left, hidden by parked flatbed and site structures $\rightarrow$ \textbf{directly impacts ego path}
    };

    \node[above=-0.1mm of output_text,xshift=-2.50cm,fill=mygreen,draw=gptblueborder,rounded corners=3pt,inner sep=2pt, text=white, font=\small] {structured JSON};

    \draw[mybluearrow, very thick] (nuscenes.east) -- (pkl.west);
    \draw[mybluearrow, very thick] (pkl.south) -- ([xshift=1.5cm]pano.north);
    \draw[mybluearrow, very thick] ([yshift=-1cm]pano.west) -- (gpt.east);
    \draw[arrow, very thick] (gpt.west) -- ([, yshift=0.0cm]output_text.east);

\draw[myblue, thick] ($(list.south east)+(0,0.1)$) -- ++(0.1,0) |- ($(list.south east)+(0.1,0)$) coordinate (bracket_bottom);
\draw[myblue, thick] (bracket_bottom) -- ([xshift=-0.1cm,yshift=-1cm]pano.west);

    \end{tikzpicture}
    \vspace{-0.4em}
    \caption{\textbf{PKL-guided dataset generation pipeline.} The process begins with nuScenes data and Occ3D visibility analysis (top left), which feeds into PKL computation for identifying planning-critical hidden agents. The selected scenes with temporal multi-camera panoramas \rnew{with agent bounding box overlay} (bottom right, t=[0,2]) are \rreplaced{processed by GPT-5 to generate structured annotations (bottom left) containing visual evidence, agent hypotheses, and occlusion analysis.}{provided to GPT-5 alongside a structured
prompt (center left) containing the ground-truth agent class and task
instructions. GPT-5 operates as a semantic annotator at this
final stage, producing structured JSON annotations (bottom left) with
visual evidence, agent hypotheses, occlusion analysis, and survey
priority.}
    }
    \label{fig:pipeline}
    \vspace{-0.6em}
\end{figure*}
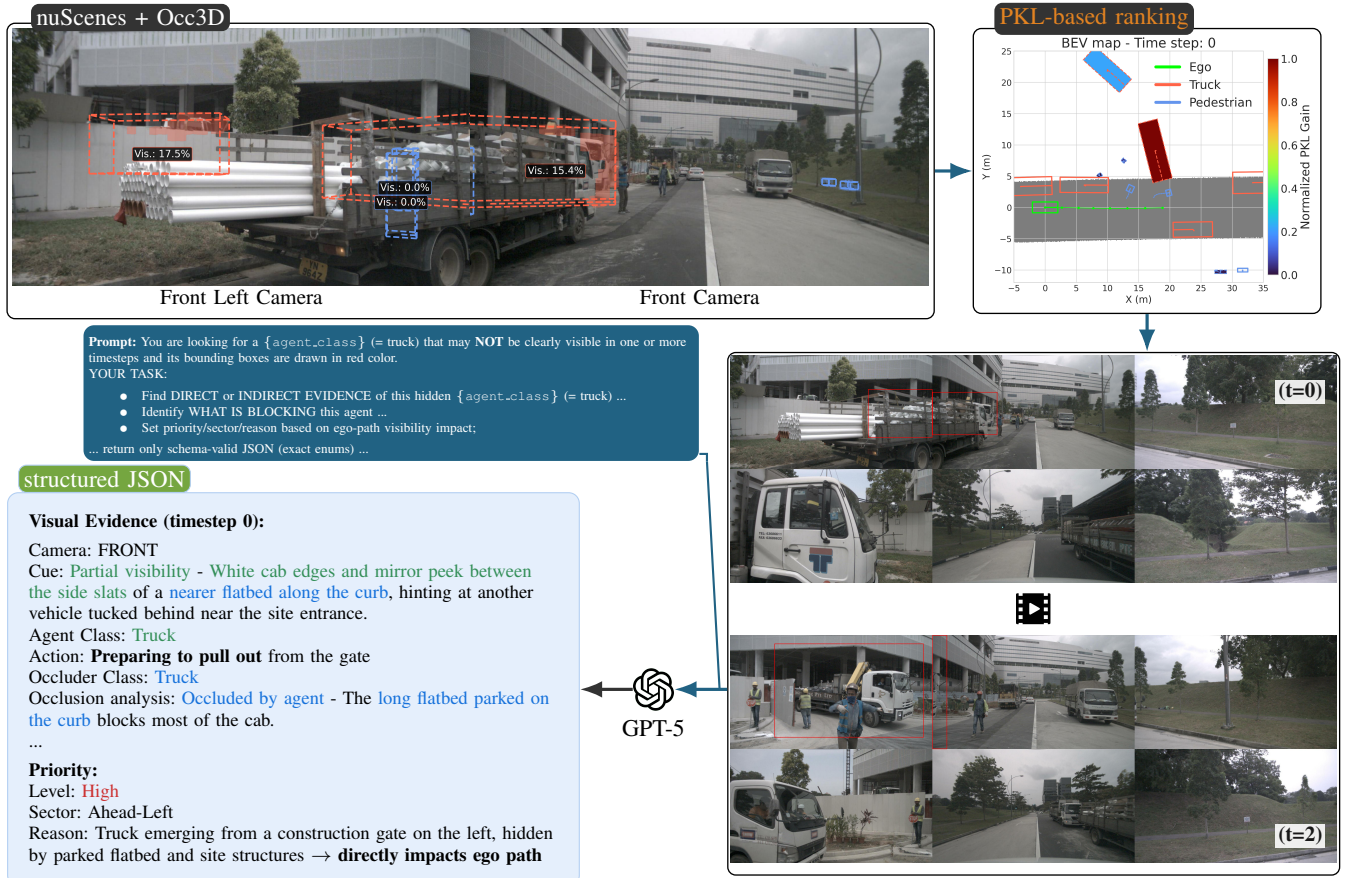
\section{Introduction}

Safe navigation requires that autonomous vehicles reason not only about what their sensors can see but also about what they cannot see. While recent advances in perception have enabled impressive object detection capabilities \cite{li2024bevformer,philion2020lift}
, a fundamental challenge remains: not all occluded regions pose equal risk to the ego vehicle's trajectory. Current approaches either treat all occlusions conservatively, leading to overly cautious behavior that disrupts traffic flow, or attempt to generatively complete the scene by predicting the contents of occluded areas, which can produce unreliable and potentially dangerous predictions \cite{yang2023parametric,tian2023occ3d}.

The gap between perception and planning becomes particularly pronounced when dealing with occluded agents. To help bridge this gap, planner-aware metrics like Planning KL-divergence (PKL) were introduced to quantify how perception errors impact downstream planning \cite{philion2020PKL}. We propose leveraging this concept for a different task: to identify \textbf{planning-critical hidden agents} whose presence would force a significant change in the ego vehicle's trajectory. However, effectively identifying these high-risk scenarios requires a deep, contextual understanding that goes beyond raw geometry. Vision-Language Models (VLMs) are well suited to this challenge, bringing semantic grounding and reasoning beyond geometry \cite{fu2023drive,chahe2025reasondrive,wang2024omnidrive}. Recent work has demonstrated VLMs' ability to handle uncertainty through techniques like conformal prediction \cite{ren2023robots} and to provide interpretable risk assessments \cite{malla2023drama}. Yet, no prior work has systematically addressed the critical challenge of determining which hidden agents actually impact safe trajectory planning using planning-aware metrics and semantic reasoning. In this paper, we present the first comprehensive framework for identifying and reasoning about planning-critical occluded agents in the autonomous driving domain using VLMs. We make three key contributions:

\begin{enumerate}
 \item We introduce a systematic method to \textbf{identify and rank planning-critical occluded agents} by applying Planning KL-Divergence (PKL) analysis to the nuScenes dataset \cite{caesar2020nuscenes}, creating a targeted collection of high-impact scenarios.
 \item We construct a new benchmark for occlusion reasoning by using GPT-5 \cite{openai2025gpt5systemcard} to generate \textbf{structured annotations} that detail the semantic context, likely agent hypotheses, and spatial reasoning for these critical events.
 \item We fine-tune and benchmark a diverse set of VLMs on this dataset, demonstrating that training on \textbf{planning-aware data} significantly improves their ability to identify and reason about high-risk occluded agents across all model scales.
\end{enumerate}

This entire data generation and annotation pipeline is illustrated in Figure~\ref{fig:pipeline}. Our work bridges the critical gap between perception uncertainty and planning safety by enabling semantically-grounded and planning-aware reasoning about occluded agents in complex urban environments.

\section{Related Work}
\label{sec:related_work}

Our research is positioned at the intersection of three key areas in autonomous driving: occlusion handling, the application of VLMs to driving scenarios, and risk-aware motion planning. \subsection{Occlusion Handling in Autonomous Driving}
Reasoning about occluded regions is a long-standing and critical challenge for ensuring the safety of autonomous vehicles (AVs). Traditional approaches often adopt a conservative stance, treating any occluded space as potentially occupied by a hidden obstacle. These methods, frequently employing set-based prediction and reachability analysis, guarantee safety by planning trajectories that avoid potential collisions under worst-case assumptions \cite{Koschi2021, Orzechowski2018}. While provably safe, this often leads to overly cautious or ``freezing" behaviors, hindering traffic efficiency, especially in dense urban environments \cite{narksri2022occlusion}. Other methods have explored Partially Observable Markov Decision Processes (POMDPs) to model the uncertainty associated with occlusions and plan maneuvers that balance safety with progress, such as ``creeping" forward to gain more visibility \cite{Hubmann2019, Wray2021}. However, these methods often lack the semantic understanding to differentiate risks based on context, thereby ignoring scene semantics and treating all blind spots equivalently.

More recently, with the rise of deep learning, Bird's-Eye-View (BEV) perception models have become a popular paradigm for representing the 3D world from 2D camera inputs \cite{li2024bevformer, philion2020lift}. Several works have attempted to address occlusions by predicting occupancy grids, effectively guessing the state of unseen areas \cite{tian2023occ3d, yang2023parametric, Zhu2023}. While these methods can reconstruct detailed scene geometry, they often struggle with a critical limitation: they treat all occlusions equally and may generate predictions that are not dynamically plausible or critical to the AV's immediate plan.
Recent works such as CorrBEV have begun to leverage language prototypes as a form of prior knowledge to enhance the features of partially occluded objects \cite{xue2025corrbev}. This emerging use of semantic, language-grounded priors to solve core perception challenges motivates our deeper exploration into the role of large-scale VLMs, which we discuss next.

\subsection{VLMs for Driving}
Recent work adapts VLMs to driving along three fronts. \emph{BEV-centric interfaces} treat the map as an image for language-grounded reasoning, e.g., Talk2BEV \cite{choudhary2024talk2bev} and BEVDriver \cite{winter2025bevdriver}. \emph{3D-aware/agentic systems} fuse multi-view perception with language models or unified world models for grounding and closed-loop control, including OmniDrive \cite{wang2024omnidrive} and NuGrounding \cite{li2025nugrounding}. \emph{Structured reasoning} frames the driving stack as a causal graph, as in DriveLM \cite{sima2024drivelm} and ELM \cite{zhou2024elm}. While these methods target general scene understanding, BEV QA, or end-to-end control, none explicitly focus on identifying \emph{which occluded agents are most critical to the immediate plan}. Our work fills this gap by coupling PKL-guided selection with structured VLM outputs for occlusion reasoning.

\subsection{Planner-Aware Metric}
To bridge the perception-planning gap, a new class of \textbf{planner-aware metrics} has been proposed~\cite{philion2020PKL, guo2020efficacy,ivanovic2022injecting}.
Our methodology is grounded in an information-theoretic approach to risk assessment. We adopt the Planning KL-Divergence (PKL) metric \cite{philion2020PKL}, originally proposed to evaluate the real-world impact of perception errors on a planner. PKL measures the divergence between a trajectory plan made with imperfect perception and an ideal plan made with ground-truth information.

While PKL was designed for evaluating perception systems, we repurpose it as a powerful tool for data curation and annotation. By measuring the change in the ego-vehicle's trajectory plan with and without knowledge of a specific hidden agent, we can quantify that agent's ``importance." This allows us to create a dataset that is not just about any occluded agent, but specifically about the \textit{most planning-critical} ones, providing a focused and challenging benchmark for occlusion reasoning.

\section{Method}
\label{s:method}

We present a systematic approach to identify and annotate planning-critical occluded agents in autonomous driving scenarios. Our method combines visibility analysis, planning-based importance ranking, and structured annotation generation to create a comprehensive dataset for training VLMs.

\subsection{Problem Formulation}
\label{ss:formulation}

Let $\mathcal{A} = \{a_1, ..., a_N\}$ denote the set of all agents in a driving scene at time $t$. We partition this set into two subsets: visible agents $\mathcal{V} \subseteq \mathcal{A}$ detected by the ego vehicle's perception system and hidden agents $\mathcal{H} \subseteq \mathcal{A}$ not detected due to occlusions or sensor limitations, such that $\mathcal{A} = \mathcal{V} \cup \mathcal{H}$.

Given a driving scene with visible agents $\mathcal{V}$ and a $T$-second window of multi-camera images, our goal is to train a VLM to generate a rich, structured description for the most planning-critical hidden agents. The model outputs a comprehensive set of attributes, covering identification, contextual analysis, and risk assessment:

\begin{itemize}
 \item \textbf{Agent Identification}: The model predicts the \textbf{agent's class} (e.g., car, pedestrian) and provides a textual description of the \textbf{visual cue} supporting its hypothesis.
 \item \textbf{Contextual Analysis}: It identifies the \textbf{occluding object}, categorizes the \textbf{occlusion type}, and localizes the event to a specific \textbf{camera view}.
 \item \textbf{Risk Assessment}: The model predicts the hidden agent's \textbf{action hypothesis}, assigns a \textbf{priority level} (high, medium, or low), and recommends a \textbf{Priority Sector} on the BEV map (e.g., Ahead-Left).
\end{itemize}
This detailed output provides an interpretable assessment of the latent risks within a scene.

\subsection{Planning-Based Importance Ranking}
\label{ss:importance}

To identify which hidden agents are planning-critical, we employ Planning KL-divergence (PKL) \cite{philion2020PKL}. The PKL measures how much the ego vehicle's planned trajectory distribution changes when given different sets of object detections:

\begin{equation}
\label{eq:pkl}
\small{
\text{PKL}(\mathcal{H})=\sum_{0<\Delta\leq T}D_{\mathrm{KL}}\!\left(
p_\theta(x_{t+\Delta}\!\mid\!\mathcal{A}_{\le t})\,\middle\|\,p_\theta(x_{t+\Delta}\!\mid\!\mathcal{V}_{\le t})
\right)
}
\end{equation}
where $T$ is the planning horizon, $p_\theta$ is the learned trajectory distribution, $x_{t+\Delta}$ is the ego vehicle's poses, $\mathcal{A}_{\leq t}$ represents the set of all agents up to time $t$, $\mathcal{V}_{\leq t}$ represents the visible agent set up to time $t$, and $D_{KL}$ denotes the Kullback-Leibler divergence.

We calculate the individual contribution, or \textbf{planning gain}, of each hidden agent $h_i \in \mathcal{H}$ by measuring the reduction in PKL if that agent were to become visible:
\begin{equation}
\label{eq:pkl-gain}
\small{
 \text{Gain}(h_i) = \text{PKL}(\mathcal{V}) - \text{PKL}(\mathcal{V} \cup \{h_i\})}
\end{equation}

This gain quantifies how much revealing agent $h_i$ would reduce the divergence between planned trajectories (i.e., the planning benefit of observing $h_i$). We use this value to rank all hidden agents in a scene, creating a ground truth importance score.

\subsection{PKL-Guided Dataset Generation Pipeline}
\label{subsec:pipeline}

Our dataset creation pipeline, illustrated in Figure~\ref{fig:pipeline}, consists of three main stages that integrate PKL-based importance ranking with VLM annotation.

\subsubsection{Hidden Agent Extraction}
We process the nuScenes dataset \cite{caesar2020nuscenes} using the Occ3D framework \cite{tian2023occ3d} to compute visibility scores for all annotated agents. For each agent $a_i$, we compute its visibility score:
\begin{equation}
\small{
 v_i = \frac{A_{\text{visible}}^i}{A_{\text{bbox}}^i}}
\end{equation}
where $A_{\text{visible}}^i$ is the area of the agent's visible voxels splatted on the camera plane, and $A_{\text{bbox}}^i$ is the area of the agent's projected 3D bounding box. We classify agents with $v_i < \tau$ as hidden agents, where $\tau$ is a visibility threshold. In Figure~\ref{fig:pipeline}'s example, the truck behind the flatbed has $v_i=0.15$, clearly qualifying as a hidden agent.

\subsubsection{Planning-Critical Agent Selection}

For each scene, we compute PKL gains for all hidden agents using Equation~\ref{eq:pkl-gain}. The BEV visualization in Figure~\ref{fig:pipeline} (top right) shows the PKL-based importance ranking, where warmer colors indicate higher planning impact. Among the multiple hidden agents in this scene, the truck preparing to pull out from the construction gate (shown in red) has the highest PKL gain, indicating it would most significantly impact the ego vehicle's trajectory. We select scenes containing at least one hidden agent with substantial planning impact, ensuring our dataset focuses on safety-critical scenarios rather than routine occlusions.

\begin{table*}[hb]
\vspace{-10pt}
\caption{Zero-shot performance of base models. Avg. action is the mean of BLEU, ROUGE-L, and Embedding Sim. Bold = best, underline = second best.}
\label{tab:base_models}
\setlength{\tabcolsep}{4pt}
\centering
\tiny
\begin{tabular}{lcc|cc|cc|cc|cc|cc|cc|ccc|ccc}
\toprule
 & \multicolumn{2}{c}{agent class} & \multicolumn{2}{c}{camera view} & \multicolumn{2}{c}{visual cue} & \multicolumn{2}{c}{occlusion type} & \multicolumn{2}{c}{occluder class} & \multicolumn{2}{c}{priority} & \multicolumn{2}{c}{sector} & \multicolumn{3}{c}{action} & \multicolumn{3}{c}{overall} \\
Model & acc. & f1 & acc. & f1 & acc. & f1 & acc. & f1 & acc. & f1 & acc. & f1 & acc. & f1 & bleu & rouge & sim. & avg. acc. & avg. f1 & avg. act\\
\midrule
Qwen-3B & 0.328 & 0.375 & 0.280 & 0.262 & 0.092 & 0.022 & 0.496 & 0.329 & 0.220 & 0.151 & 0.088 & 0.054 & 0.128 & 0.089 & 0.000 & 0.000 & 0.249 & 0.233 & 0.183 & 0.083 \\
Qwen-7B & 0.432 & 0.501 & 0.252 & 0.229 & 0.124 & 0.093 & 0.508 & 0.374 & 0.248 & 0.204 & 0.192 & 0.074 & 0.352 & 0.324 & \underline{0.000} & 0.003 & 0.253 & 0.301 & 0.257 & 0.085 \\
Qwen-32B & 0.512 & 0.558 & \underline{0.296} & \underline{0.307} & 0.492 & 0.508 & 0.612 & 0.582 & 0.292 & 0.274 & 0.252 & 0.181 & 0.300 & 0.308 & 0.000 & 0.009 & 0.270 & 0.394 & 0.388 & 0.093 \\
Qwen-72B & 0.516 & 0.565 & 0.204 & 0.125 & 0.640 & 0.587 & 0.624 & 0.583 & 0.348 & \underline{0.320} & 0.364 & 0.323 & \textbf{0.384} & 0.338 & 0.000 & 0.012 & 0.266 & \textbf{0.440} & \textbf{0.406} & 0.093 \\
\midrule
Intern-1B & \underline{0.644} & \underline{0.627} & 0.000 & 0.000 & 0.000 & 0.000 & 0.000 & 0.000 & 0.000 & 0.000 & 0.000 & 0.000 & 0.000 & 0.000 & 0.000 & 0.001 & 0.246 & 0.092 & 0.090 & 0.082 \\
Intern-2B & 0.324 & 0.424 & 0.212 & 0.190 & \textbf{0.800} & \textbf{0.726} & 0.496 & 0.329 & 0.248 & 0.169 & 0.184 & 0.057 & 0.356 & \underline{0.342} & 0.000 & 0.012 & 0.246 & 0.374 & 0.319 & 0.086 \\
Intern-4B & \textbf{0.648} & \textbf{0.639} & 0.088 & 0.108 & 0.264 & 0.342 & 0.264 & 0.280 & 0.084 & 0.084 & 0.168 & 0.202 & 0.112 & 0.131 & 0.000 & 0.009 & 0.266 & 0.233 & 0.255 & 0.092 \\
Intern-8B & 0.496 & 0.544 & 0.240 & 0.227 & 0.628 & 0.563 & \underline{0.648} & \underline{0.617} & 0.260 & 0.224 & \textbf{0.548} & 0.388 & 0.248 & 0.201 & 0.000 & \underline{0.013} & 0.271 & \underline{0.438} & \underline{0.395} & 0.095 \\
Intern-14B & 0.328 & 0.416 & 0.168 & 0.156 & 0.648 & 0.579 & \textbf{0.692} & \textbf{0.658} & \textbf{0.392} & \textbf{0.340} & 0.184 & 0.057 & 0.208 & 0.135 & 0.000 & 0.008 & \underline{0.277} & 0.374 & 0.334 & 0.095 \\
Intern-GPT & 0.708 & 0.659 & 0.188 & 0.168 & 0.080 & 0.012 & 0.436 & 0.295 & 0.120 & 0.104 & 0.204 & 0.103 & 0.204 & 0.131 & 0.000 & 0.016 & 0.273 & 0.277 & 0.210 & 0.096 \\
Intern-38B & 0.220 & 0.295 & 0.096 & 0.043 & \underline{0.712} & \underline{0.605} & 0.608 & 0.551 & \underline{0.368} & 0.316 & 0.192 & 0.080 & 0.152 & 0.083 & 0.000 & 0.010 & \textbf{0.278} & 0.335 & 0.282 & \underline{0.096} \\
\midrule
Gemma-4B & 0.504 & 0.558 & 0.268 & 0.237 & 0.088 & 0.014 & 0.436 & 0.333 & 0.168 & 0.160 & 0.480 & \underline{0.394} & \underline{0.380} & \textbf{0.358} & \textbf{0.008} & 0.011 & 0.224 & 0.332 & 0.293 & 0.081 \\
Gemma-12B & 0.300 & 0.391 & 0.280 & 0.263 & 0.076 & 0.029 & 0.464 & 0.425 & 0.228 & 0.221 & 0.344 & 0.317 & 0.176 & 0.122 & 0.000 & 0.007 & 0.242 & 0.267 & 0.253 & 0.083 \\
Gemma-27B & 0.484 & 0.527 & 0.276 & 0.264 & 0.108 & 0.053 & 0.536 & 0.473 & 0.324 & 0.284 & 0.484 & \textbf{0.416} & 0.240 & 0.226 & 0.000 & 0.001 & 0.245 & 0.350 & 0.320 & 0.082 \\
\midrule
MiniCPM & 0.468 & 0.525 & 0.144 & 0.102 & 0.208 & 0.262 & 0.568 & 0.509 & 0.336 & 0.288 & \textbf{0.548} & 0.388 & 0.216 & 0.201 & 0.000 & 0.000 & 0.247 & 0.355 & 0.325 & 0.082 \\
\midrule

DriveMM & 0.004 & 0.008 & 0.000 & 0.000 & 0.004 & 0.008 & 0.000 & 0.000 & 0.000 & 0.000 & 0.000 & 0.000 & 0.000 & 0.000 & 0.000 & 0.000 & 0.275 & 0.001 & 0.002 & 0.092 \\
DriveLM & 0.148 & 0.236 & 0.180 & 0.187 & 0.536 & 0.558 & 0.004 & 0.008 & 0.000 & 0.000 & 0.184 & 0.057 & 0.132 & 0.156 & 0.000 & \textbf{0.022} & 0.271 & 0.169 & 0.172 & \textbf{0.098} \\
DriveGPT4 & 0.000 & 0.000 & \textbf{0.460} & \textbf{0.311} & \underline{0.712} & 0.592 & 0.424 & 0.266 & 0.028 & 0.037 & 0.180 & 0.056 & 0.308 & 0.276 & 0.000 & 0.000 & 0.000 & 0.302 & 0.220 & 0.000 \\
\bottomrule
\end{tabular}

\end{table*}

\subsubsection{Structured Annotation Generation}
\label{sudsec:gpt-gen}
The key innovation of our approach lies in generating rich semantic annotations for planning-critical hidden agents. As shown in Figure~\ref{fig:pipeline} (bottom), we provide GPT-5 with temporal multi-camera panoramas spanning $T$ seconds and the target hidden agent's class and bounding box to generate structured descriptions.GPT-5 generates structured JSON annotations containing:
\begin{itemize}
 \item \textbf{Visual Cue}: Evidence indicating the hidden agent $\in$ \{partial visibility, indirect motion responses, temporal traces, light signatures\}, along with a textual description
 \item \textbf{Camera}: View where evidence is observed (e.g., ``FRONT'')
 \item \textbf{Agent Class}: Hidden agent class $\in$ \{car, truck, trailer, bus, motorcycle, pedestrian, bicycle, unknown\}
 \item \textbf{Action Hypothesis}: Predicted behavior (e.g., ``Preparing to pull out from the gate'')
 \item \textbf{Occluder Class}: Primary occluding object's nuScenes class (e.g., ``Truck'')
 \item \textbf{Occlusion Type}: Occlusion pattern $\in$ \{occluded by agent, occluded by landmark, weather, lighting, other\}, along with a textual description (e.g., ``The long flatbed parked on the curb blocks most of the cab'')
 \item \textbf{Priority}: Including Priority Level $\in$ \{High, Medium, Low\} and Sector in ego's BEV $\in$ \{Ahead-Left, Ahead, Ahead-Right, Left, Right, Back-Left, Back, Back-Right\}
 \item \textbf{Reasoning}: Explanation for the risk assessment
\end{itemize}

We iteratively refined the annotation process by optimizing prompts~\cite{openai2025gpt5systemcard} to emphasize indirect visual cues over direct visibility, standardizing coordinate-system definitions for consistent localization, issuing detailed classification guidelines to reduce ambiguity. We manually auditeda subset of samples to ensure annotation quality and consistency.
\definecolor{lightgreen}{HTML}{90EE90}
\definecolor{lightred}{HTML}{FFB6C1}
\begin{figure}[h]
    \centering

    \subcaptionbox{Panoramic \rreplaced{input}{front} view at timestep 2}[1\columnwidth]{%
        \includegraphics[width=\columnwidth, trim={0 900bp 0 0}, clip]{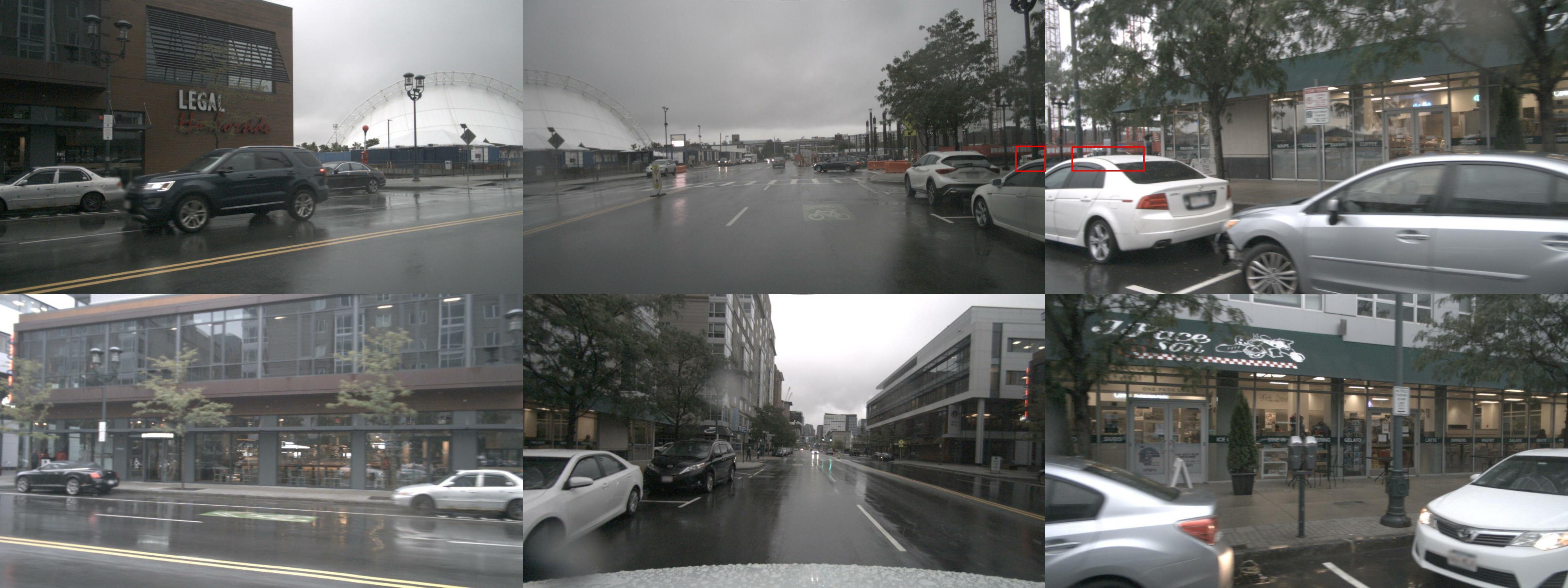}
    }
    \vspace{1pt} 


    \begin{minipage}[h]{0.49\columnwidth} 
        \centering
        \vspace{0pt}
        \subcaptionbox{BEV map}[1\linewidth]{%
            \includegraphics[width=\linewidth]{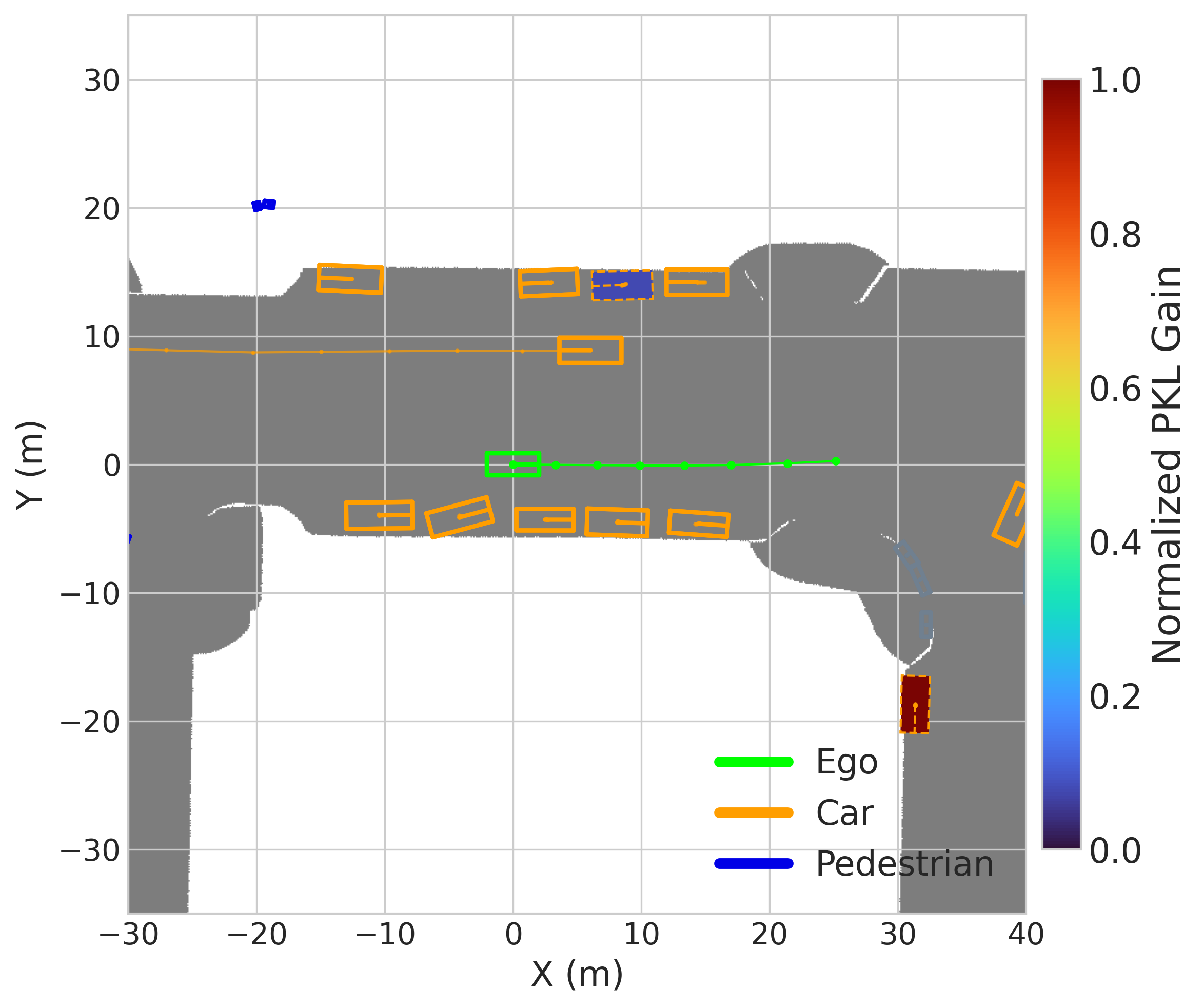}
        }
    \end{minipage}%
    \hfill 
\begin{minipage}[h]{0.5\columnwidth} 
    \centering
    \vspace{0pt}
    \subcaptionbox{Model predictions}[1\linewidth]{%
        \tiny 
        \setlength{\tabcolsep}{2pt} 
        \renewcommand{\arraystretch}{1.2} 
        \begin{tabular}{@{} l c c c c @{}}
            \toprule
             & \multicolumn{3}{c}{\textbf{InternVL3.5-Ins}} & \\ 
            \cmidrule(lr){2-4} 
            \textbf{Field} & \textbf{4B} & \textbf{GPT} & \textbf{38B} & \textbf{SFT}\\
            \midrule
            \texttt{class} & \cellcolor{lightgreen}car & \cellcolor{lightgreen}car & \cellcolor{lightred}pedestrian & \cellcolor{lightgreen}car \\
            \texttt{action} & \cellcolor{lightred}turn\_left & \cellcolor{lightred}turn\_left & \cellcolor{lightred}brake & \cellcolor{lightgreen}parked \\
            \texttt{cam\_view} & \cellcolor{lightred}F\_LEFT & \cellcolor{lightred}F\_LEFT & \cellcolor{lightred}F\_LEFT & \cellcolor{lightgreen}F\_RIGHT \\
            \texttt{vis\_cue} & \cellcolor{lightgreen}partial & \cellcolor{lightred}indirect & \cellcolor{lightgreen}partial & \cellcolor{lightgreen}partial \\
            \texttt{occ\_type} & \cellcolor{lightred}landmark & \cellcolor{lightred}landmark & \cellcolor{lightgreen}agent & \cellcolor{lightgreen}agent \\
            \texttt{occ\_class} & \cellcolor{lightred}manmade & \cellcolor{lightred}manmade & \cellcolor{lightgreen}car & \cellcolor{lightgreen}car \\
            \texttt{priority} & \cellcolor{lightred}high & \cellcolor{lightred}high & \cellcolor{lightred}high & \cellcolor{lightgreen}medium \\
            \texttt{sector} & \cellcolor{lightred}A\_Left & \cellcolor{lightred}A\_Left & \cellcolor{lightred}A\_Left & \cellcolor{lightgreen}A\_Right \\
            \bottomrule
        \end{tabular}
    }
\end{minipage}

    \caption{A rainy intersection scene with an occluded car. Fine-tuning (SFT) the Intern\rremoved{VL} models (from 4B to 38B) results in more plausible predictions compared to their zero-shot versions. \rreplaced{Green and red highlights indicate correct and incorrect predictions, respectively.}{Green = correct, red = incorrect.}
    \vspace{-15pt}}
    \label{fig:intrnvl}
\end{figure}

\section{Experiments}
\label{sec:experiments}
We evaluate our approach on the nuScenes dataset, demonstrating that VLMs can effectively identify planning-critical occluded agents when fine-tuned on PKL-guided data. Our experiments show significant improvements across all model families, with fine-tuned smaller models outperforming larger zero-shot counterparts.

\subsection{Experimental Setup}
\label{subsec:setup}

\subsubsection{Dataset}
We construct a comprehensive dataset from nuScenes~\cite{caesar2020nuscenes} using the PKL-guided pipeline described in Section~\ref{subsec:pipeline}. To compute PKL scores, we train the planner following the methodology in~\cite{philion2020PKL} on the nuScenes training split, producing trajectory predictions over a
4\,s horizon. PKL is then computed over $T{=}2$\,s windows within each scene. We set the visibility threshold $\tau{=}0.4$. Both the chunking window and threshold choices are validated in Section~\ref{sec:pkl-sensitivity}. We select scenes containing at least one hidden agent with substantial planning impact by sorting all scenes by their $\text{PKL}(\mathcal{V})$ values and choosing the top 1,000 from the nuScenes training set and top 250 from the nuScenes validation set, yielding diverse occlusion scenarios from urban driving. This selection ensures our dataset focuses on safety-critical situations where hidden agents would significantly alter the ego vehicle's trajectory if visible.
For each selected scene, we use GPT-5 to generate structured JSON annotations as described in Section~\ref{sudsec:gpt-gen}. Our final dataset comprises 1,250 scene windows(1,000 training, 250 validation), resulting in 22,500 camera frames (18,000 training, 4,500 validation). Each windowspans $T{=}2$ seconds with 3 frames sampled at 1-second intervals across 6 cameras in a 3$\times$2 grid configuration. This yields 3,750 structured JSON annotations (3,000 training, 750 validation) generated by GPT-5, with one annotation per timestep.

\begin{table*}[h]
\caption{Supervised fine-tuned model performance. Values in parentheses show improvement over base models. Avg. action is the mean of BLEU, ROUGE-L, and Embedding Sim. Bold = best, underline = second best.}
\label{tab:sft_models}
\setlength{\tabcolsep}{4pt}
\centering
\tiny
\begin{tabular}{lcc|cc|cc|cc|cc|cc|cc|ccc|ccc}
\toprule
 & \multicolumn{2}{c}{agent class} & \multicolumn{2}{c}{camera view} & \multicolumn{2}{c}{visual cue} & \multicolumn{2}{c}{occlusion type} & \multicolumn{2}{c}{occluder class} & \multicolumn{2}{c}{priority} & \multicolumn{2}{c}{sector} & \multicolumn{3}{c}{action} & \multicolumn{3}{c}{overall} \\
Model & acc. & f1 & acc. & f1 & acc. & f1 & acc. & f1 & acc. & f1 & acc. & f1 & acc. & f1 & bleu & rouge & sim. & avg. acc. & avg. f1 & act avg. \\
\midrule
Qwen-3B & 0.708 & 0.683 & 0.196 & 0.184 & 0.812 & 0.741 & 0.628 & 0.601 & 0.352 & 0.331 & 0.544 & 0.387 & 0.104 & 0.125 & 0.319 & 0.364 & 0.546 & 0.478 (+.24) & 0.436 (+.25) & 0.410 (+.33) \\
Qwen-7B & 0.756 & 0.741 & 0.340 & 0.322 & 0.804 & 0.728 & 0.624 & 0.600 & 0.412 & 0.370 & 0.548 & 0.494 & 0.320 & 0.342 & 0.329 & 0.407 & 0.569 & 0.543 (+.24) & 0.514 (+.26) & 0.435 (+.35) \\
Qwen-32B & 0.660 & 0.667 & 0.272 & 0.273 & 0.744 & 0.654 & 0.620 & 0.603 & 0.412 & 0.385 & 0.468 & 0.447 & 0.204 & 0.228 & 0.320 & 0.372 & 0.547 & 0.483 (+.09) & 0.465 (+.08) & 0.413 (+.32) \\
\midrule
Intern-1B & 0.756 & 0.692 & 0.452 & 0.304 & 0.804 & 0.728 & 0.488 & 0.361 & 0.288 & 0.187 & 0.548 & 0.388 & 0.476 & 0.327 & 0.319 & 0.377 & 0.553 & 0.545 (+.45) & 0.427 (+.34) & 0.416 (+.33) \\
Intern-2B & 0.768 & 0.704 & 0.464 & 0.326 & 0.756 & 0.674 & 0.576 & 0.521 & 0.392 & 0.332 & 0.548 & 0.388 & 0.508 & 0.442 & 0.329 & 0.390 & 0.566 & 0.573 (+.20) & 0.484 (+.16) & 0.428 (+.34) \\
Intern-4B & 0.752 & 0.684 & 0.428 & 0.335 & 0.800 & 0.724 & 0.608 & 0.591 & 0.412 & 0.373 & 0.480 & 0.430 & 0.476 & 0.407 & 0.316 & 0.427 & 0.584 & 0.565 (+.33) & 0.506 (+.25) & \underline{0.443 (+.35)} \\
Intern-8B & 0.752 & 0.684 & 0.396 & 0.336 & 0.792 & 0.716 & 0.596 & 0.555 & 0.364 & 0.300 & 0.564 & 0.467 & 0.492 & 0.415 & 0.319 & 0.416 & 0.585 & 0.565 (+.13) & 0.496 (+.10) & 0.440 (+.35) \\
Intern-14B & 0.740 & 0.657 & 0.468 & 0.384 & 0.796 & 0.720 & 0.636 & 0.616 & 0.464 & 0.425 & 0.556 & 0.498 & 0.496 & 0.435 & 0.334 & 0.422 & 0.582 & \textbf{0.594 (+.22)} & \underline{0.534 (+.20)} & \textbf{0.446 (+.35)} \\
Intern-GPT & 0.736 & 0.643 & 0.488 & 0.349 & 0.712 & 0.592 & 0.528 & 0.407 & 0.308 & 0.177 & 0.532 & 0.387 & 0.468 & 0.329 & 0.320 & 0.423 & 0.580 & 0.539 (+.26) & 0.412 (+.20) & 0.441 (+.34) \\
\midrule
Gemma-4B & 0.740 & 0.670 & 0.468 & 0.299 & 0.804 & 0.730 & 0.620 & 0.612 & 0.340 & 0.301 & 0.408 & 0.377 & 0.228 & 0.196 & 0.313 & 0.378 & 0.546 & 0.515 (+.18) & 0.455 (+.16) & 0.412 (+.33) \\
Gemma-12B & 0.740 & 0.682 & 0.460 & 0.347 & 0.800 & 0.726 & 0.676 & 0.665 & 0.428 & 0.378 & 0.576 & 0.512 & 0.440 & 0.386 & 0.305 & 0.395 & 0.570 & \underline{0.589 (+.32)} & 0.528 (+.28) & 0.423 (+.34) \\
Gemma-27B & 0.728 & 0.704 & 0.364 & 0.368 & 0.804 & 0.728 & 0.652 & 0.653 & 0.440 & 0.425 & 0.520 & 0.494 & 0.404 & 0.390 & 0.295 & 0.379 & 0.556 & 0.559 (+.21) & \textbf{0.538 (+.22)} & 0.410 (+.33) \\
\midrule
MiniCPM & 0.756 & 0.671 & 0.416 & 0.339 & 0.792 & 0.716 & 0.612 & 0.585 & 0.420 & 0.366 & 0.552 & 0.403 & 0.484 & 0.396 & 0.315 & 0.414 & 0.589 & 0.576 (+.22) & 0.497 (+.17) & 0.439 (+.36) \\
\midrule
DriveLM & 0.656 & 0.602 & 0.428 & 0.342 & 0.652 & 0.581 & 0.488 & 0.481 & 0.224 & 0.198 & 0.520 & 0.396 & 0.432 & 0.297 & 0.284 & 0.355 & 0.534 & 0.486 (+.32) & 0.414 (+.24) & 0.391 (+.29)\\
\bottomrule
\end{tabular}
\vspace{-15pt}
\end{table*}

\subsubsection{Evaluation Metrics}
We evaluate model performance against the ground truth annotations across the structured fields described in our method: Visual Cue, Camera, Agent Class, Action Hypothesis, Occluder Class, Occlusion Type, Priority Level, and Sector.
To measure performance, we use the following metrics:

 \textbf{Classification Metrics}: For all categorical fields (Agent Class, Camera, Occluder Class, Occlusion Type, Priority Level, and Sector), we compute Accuracy and F1 scores.

 \textbf{Text Similarity Metrics}: For the Action Hypothesis field, we reportBLEU, ROUGE-L, and embedding similarity~\cite{reimers-2019-sentence-bert}.

 \textbf{Overall Performance}: We report average accuracy and average F1 across all categorical fields, and average action score (mean of BLEU, ROUGE-L, and embedding similarity). These provide complementary views of classification consistency and text generation quality.
\definecolor{lightgreen}{HTML}{90EE90}
\definecolor{lightred}{HTML}{FFB6C1}
\begin{figure}[hb]
    \centering

    \subcaptionbox{Panoramic \rreplaced{input}{front} view at timestep 0}[\columnwidth]{%
        \includegraphics[width=\columnwidth, trim={0 900bp 0 0}, clip]{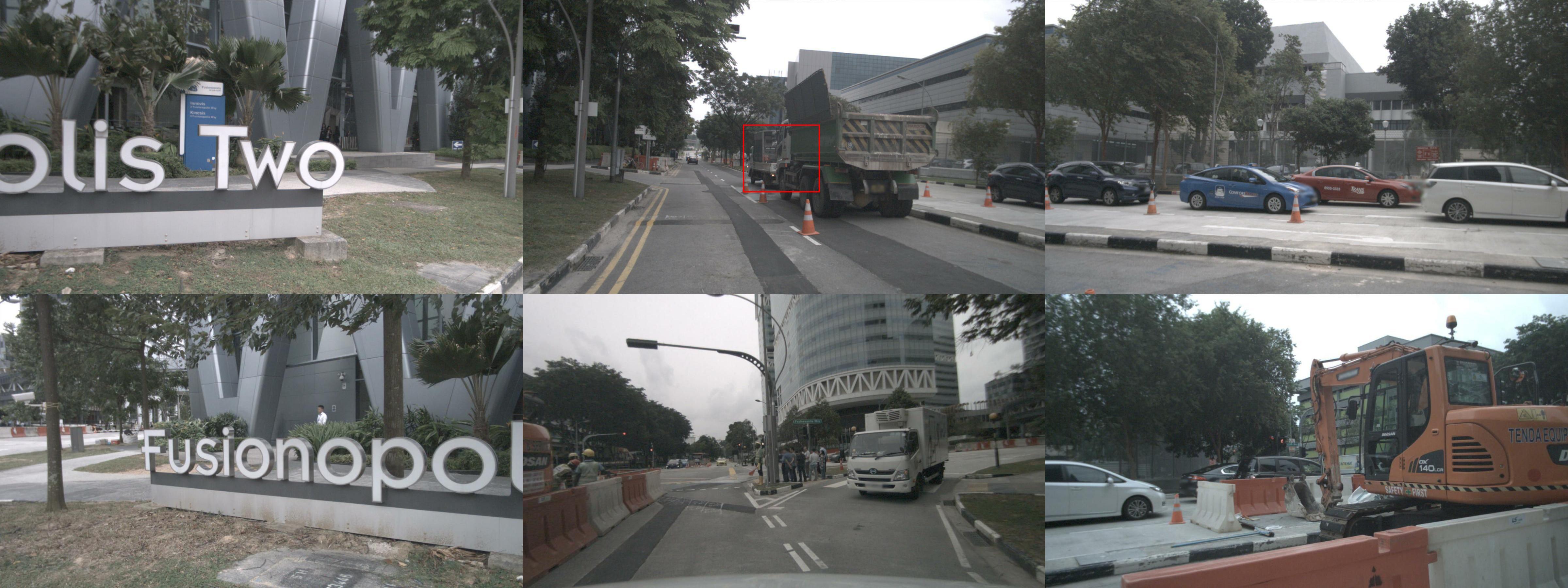}
    }
    \vspace{1pt} 


    \begin{minipage}[h]{0.45\columnwidth} 
        \centering
        \vspace{0pt}
        \subcaptionbox{BEV map}[1\linewidth]{%
            \includegraphics[width=\linewidth]{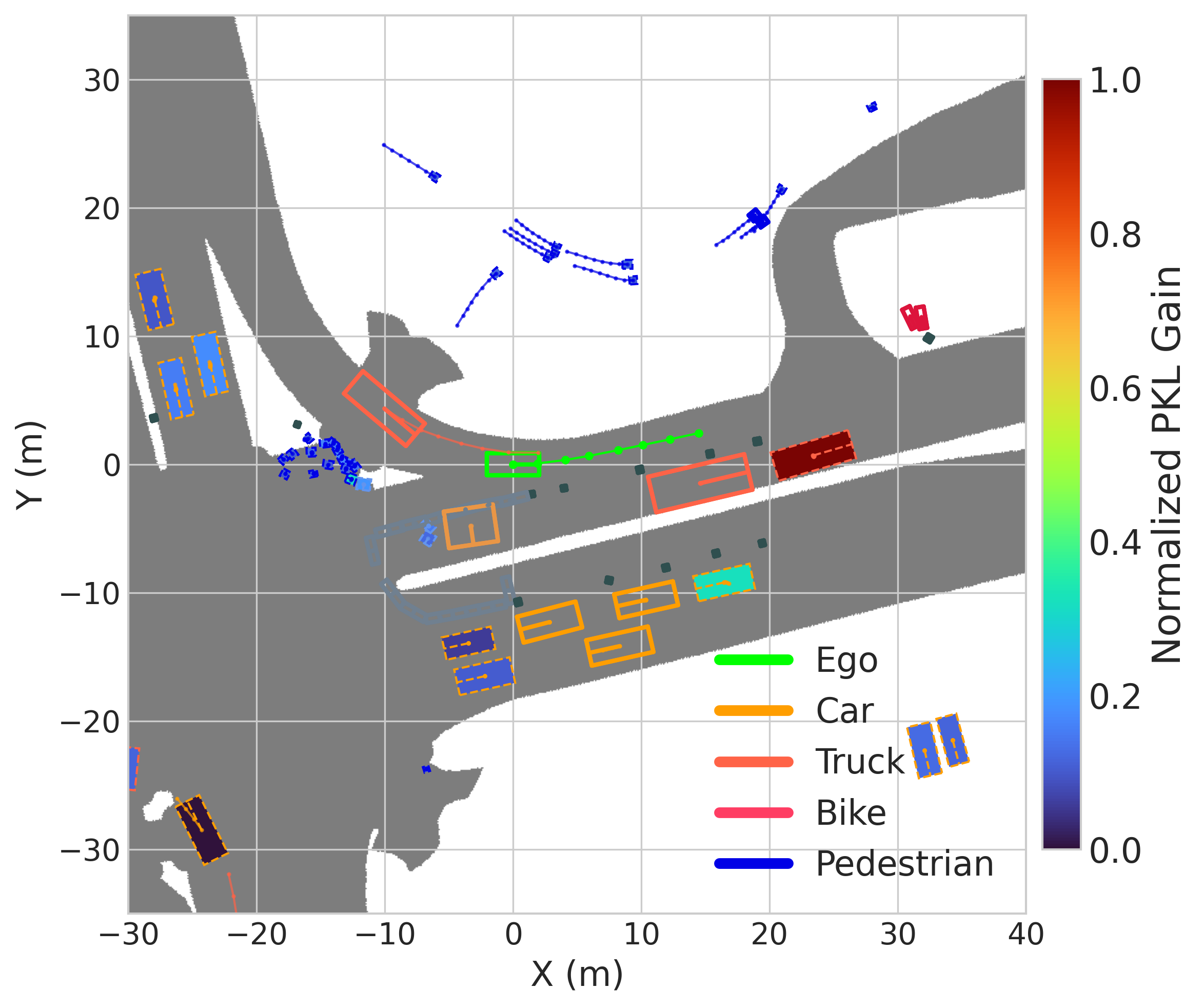}
        }
    \end{minipage}%
    \hfill 
\begin{minipage}[h]{0.54\columnwidth}
    \centering
    \vspace{0pt}
    \subcaptionbox{Model predictions}[1\linewidth]{%
        \tiny 
        \setlength{\tabcolsep}{1.5pt} 
        \renewcommand{\arraystretch}{1.2} 
        \begin{tabular}{@{} l c c c c @{}}
            \toprule
             & \multicolumn{2}{c}{\textbf{Gemma-3-12B}} & \multicolumn{2}{c}{\textbf{Gemma-3-27B}} \\ 
            \cmidrule(lr){2-3} \cmidrule(lr){4-5} 
            \textbf{Field} & \textbf{Ins} & \textbf{SFT} & \textbf{Ins} & \textbf{SFT} \\
            \midrule
            \texttt{class} & \cellcolor{lightgreen}truck & \cellcolor{lightgreen}truck & \cellcolor{lightred}car & \cellcolor{lightgreen}truck \\
            \texttt{action} & \cellcolor{lightred}accelerate & \cellcolor{lightred}moving forw. & \cellcolor{lightred}accelerate & \cellcolor{lightgreen}stopped \\
            \texttt{cam\_view} & \cellcolor{lightgreen}FRONT & \cellcolor{lightgreen}FRONT & \cellcolor{lightred}F\_RIGHT & \cellcolor{lightgreen}FRONT \\
            \texttt{vis\_cue} & \cellcolor{lightred}temp. trace & \cellcolor{lightgreen}partial & \cellcolor{lightred}temp. trace & \cellcolor{lightgreen}partial \\
            \texttt{occ\_type} & \cellcolor{lightred}landmark & \cellcolor{lightgreen}agent & \cellcolor{lightgreen}agent & \cellcolor{lightgreen}agent \\
            \texttt{occ\_class} & \cellcolor{lightred}construction & \cellcolor{lightgreen}truck & \cellcolor{lightgreen}truck & \cellcolor{lightgreen}truck \\
            \texttt{priority} & \cellcolor{lightgreen}high & \cellcolor{lightgreen}high & \cellcolor{lightred}medium & \cellcolor{lightgreen}high \\
            \texttt{sector} & \cellcolor{lightred}Ahead & \cellcolor{lightred}Ahead & \cellcolor{lightgreen}A\_Right & \cellcolor{lightgreen}A\_Right \\
            \bottomrule
        \end{tabular}
    }
\end{minipage}

    \caption{An urban construction scene with a hidden truck. Fine-tuning (SFT) significantly improves the Gemma models' predictions\rremoved{ for the agent's action and priority}. \rreplaced{Green and red highlights indicate correct and incorrect predictions, respectively.}{Green = correct, red = incorrect.}
    \vspace{-10pt}}
    \label{fig:gemma}
\end{figure}
\subsubsection{Implementation Details}
All models are fine-tuned using LoRA~\cite{hu2022lora} with rank 16, targeting all linear layers including the alignment module and language model components. We employ the ms-swift framework~\cite{zhao2024msswift} with AdamW optimizer (learning rate 2e-5), batch size 4 per GPU across 4 H100 GPUs, and train for 3 epochs. For inference, we use structured JSON output generation to ensure consistent formatting across all models.

\subsection{Baseline Models}
\label{subsec:baselines}
We evaluate two categories of VLMs:

\noindent\textbf{General-Purpose VLMs:}
\begin{itemize}
 \item \textbf{Qwen2.5-VL}~\cite{bai2025qwen2}: 3B, 7B, 32B, 72B parameters.
 \item \textbf{InternVL3.5}~\cite{wang2025internvl3}: 1B, 2B, 4B, 8B, 14B, 38B, GPT-OSS-20B configurations.
 \item \textbf{Gemma-3}~\cite{team2025gemma}: 4B, 12B, 27B variants.
 \item \textbf{MiniCPM-V 4.5}~\cite{yao2024minicpm}: 7B parameter.
\end{itemize}

\noindent\textbf{Domain-Adapted Models:}
\begin{itemize}
 \item \textbf{DriveMM}~\cite{huang2024drivemm}: Enhanced LLaVA-7B with driving-specific adaptations.
 \item \textbf{DriveLM}~\cite{sima2024drivelm}: Mini-InternVL2-4B fine-tuned on DriveLM dataset~\cite{gao2024mini}.
 \item \textbf{DriveGPT4}~\cite{xu2024drivegpt4}: Mini-InternVL2-4B fine-tuned on DriveGPT4 dataset~\cite{gao2024mini}.
\end{itemize}

\subsection{Main Results}
\label{subsec:main_results}
\subsubsection{Zero-Shot Performance}
Table~\ref{tab:base_models} presents zero-shot performance of base (Instruct) and domain-adapted models. Key observations:

\textit{Scale Effects:} Larger models generally achieve better zero-shot performance within each family. Qwen-72B achieves the highest average accuracy (0.440) among open models, followed by InternVL3.5-8B (0.438).
\textit{Cross-Family Variations:} Despite size differences, performance varies significantly across model families. InternVL3.5-8B (8B parameters) matches Qwen-72B's performance, suggesting architectural differences play a crucial role.

\textit{Domain Adaptation Limitations:} Surprisingly, domain-adapted models show mixed results. While DriveLM achieves the highest average action score (0.098), DriveMM fails almost completely (0.001 avg.\ accuracy), and DriveGPT4 shows no action understanding. This suggests that general driving knowledge alone is insufficient for this task.

\subsubsection{Fine-Tuning Impact}
Table~\ref{tab:sft_models} demonstrates the transformative effect of supervised fine-tuning on our PKL-guided dataset:

\textit{Dramatic Improvements:} Fine-tuning yields substantial gains across all models, with average accuracy improvements ranging from +0.09 (Qwen-32B) to +0.45 (InternVL3.5-1B). Smaller models benefit disproportionately: InternVL3.5-1B improves from 0.092 to 0.545 in average accuracy (+493\%). Figure~\ref{fig:gemma} provides a qualitative example where fine-tuning enables correct identification of a hidden truck and its priority level, while base (instruct) models fail to capture these critical details.

\textit{Efficiency Gains:} Fine-tuned smaller models consistently outperform larger zero-shot variants. Notably, Qwen-3B-SFT (0.478 avg.\ accuracy) surpasses Qwen-72B zero-shot (0.440) while being 24$\times$ smaller, demonstrating the value of task-specific training. Figure~\ref{fig:intrnvl} illustrates this phenomenon within the InternVL3.5 family, where the fine-tuned version consistently outperforms larger base (instruct) variants.

\textit{Component-Specific Gains:} Fine-tuning benefits vary substantially across evaluation targets. Visual cue detection shows the largest gain (+950\%), followed by priority prediction (+518\%), occluder classification (+390\%), camera view (+386\%), and action prediction (+138\% in embedding similarity). This ordering suggests that fine-tuning is most impactful for skills absent in pretraining (e.g., reasoning about indirect occlusion cues and inferring the agent actions).

\subsection{Ablation Studies}
\label{subsec:ablations}
\subsubsection{PKL and Visibility Threshold Sensitivity}
\label{sec:pkl-sensitivity}

\begin{figure}[hb]
    \centering
    \begin{minipage}[b]{0.22\textwidth}
        \centering
        \includegraphics[width=\linewidth]{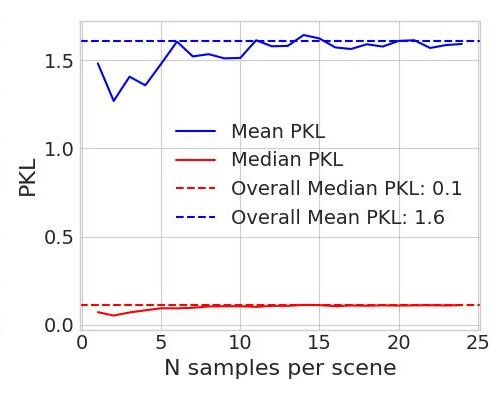}
    \end{minipage}%
    \hfill
    \begin{minipage}[b]{0.264\textwidth}
        \centering
        \includegraphics[width=\linewidth]{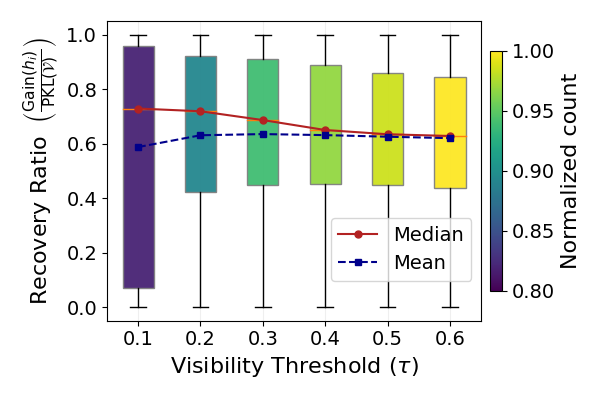}
    \end{minipage}
    \rnew{\caption{Robustness of PKL estimation and visibility 
threshold. (Left) Per-scene PKL convergence: mean and median 
stabilize after $\sim$5 samples. (Right) PKL Recovery Ratio 
across $\tau \in [0.1,0.6]$: recovery remains stable 
while lower thresholds yield fewer but more severely 
occluded agents.
}}
    \label{fig:pkl-sensitivity}
\end{figure}

We validate the robustness of two key hyperparameters in our pipeline. Figure~\ref{fig:pkl-sensitivity} (left) shows the per-scene PKL convergence analysis on the nuScenes training set. Both mean and median PKL estimates stabilize after approximately 2 seconds (5 samples per scene), confirming that our $T{=}2$\,s windows provide reliable importance estimates without requiring exhaustive sampling.

Figure~\ref{fig:pkl-sensitivity} (right) reports the PKL Recovery
Ratio ($\frac{\text{Gain}(h_i)}{\text{PKL}(\mathcal{V})}$) on the
nuScenes validation set across $\tau \in [0.1, 0.6]$. Mean and
median recovery remain stable at 0.6-0.75. However, $\tau$ also controls a trade-off:
lower values retain fewer, more severely occluded agents, while
higher values include more agents. Our
$\tau{=}0.4$ balances dataset size with occlusion severity, aligns
with the nuScenes lowest visibility bin (0--40\%), and lies within
the stable recovery region.

\subsubsection{Impact of PKL-Guided Selection}
Table~\ref{tab:pkl-abliation} validates our PKL-guided data selection strategy by comparing against random hidden agent selection. Both models were evaluated on the same 250-sample validation set from nuScenes to ensure fair comparison. For Qwen-7B, PKL-guided selection improves average accuracy from 0.405 to 0.543 (+34\%). Similarly, InternVL3.5-8B improves from 0.429 to 0.565 in accuracy (+32\%). Figure~\ref{fig:abli-night} provides a qualitative example of this performance gap; it illustrates a common failure case for models trained on data with randomly selected hidden agents, which often misidentify agent priority and wrongly point to less important hidden agents.
The consistent $\sim$30\% improvement demonstrates that focusing on planning-critical agents during training is essential. The gains in action and class metrics indicate a better semantic understanding of occlusion patterns. This validates our hypothesis that the model benefits significantly from training on examples where hidden agents substantially impact planning decisions.

\begin{table}[h]
\centering
\caption{Impact of PKL-guided data selection. Top row: Random selection baseline. Bottom row: PKL selection. Bold values indicate an improvement over the baseline (deltas in parentheses).}
\label{tab:pkl-abliation}
\scriptsize
\setlength{\tabcolsep}{3pt}
\begin{tabular}{ll lll|l}
\toprule
 & & \multicolumn{3}{c}{\textbf{Average Class}} & \textbf{Action} \\
\cmidrule(lr){3-5} \cmidrule(lr){6-6}
\textbf{Model} & \textbf{Sel.} & Acc. & Prec. & F1 & Avg. \\
\midrule
\multirow{2}{*}{Qwen-7B} & Rand. & 0.405 & 0.446 & 0.350 & 0.410 \\
 & PKL & \textbf{0.543}{\scriptsize (+.138)} & \textbf{0.533}{\scriptsize (+.087)} & \textbf{0.514}{\scriptsize (+.164)} & \textbf{0.435}{\scriptsize (+.025)} \\
\midrule
\multirow{2}{*}{Intern-8B} & Rand. & 0.429 & 0.419 & 0.382 & 0.406 \\
 & PKL & \textbf{0.565}{\scriptsize (+.136)} & \textbf{0.486}{\scriptsize (+.067)} & \textbf{0.496}{\scriptsize (+.114)} & \textbf{0.440}{\scriptsize (+.034)} \\
\bottomrule
\end{tabular}
\end{table}
\vspace{-15pt}
\definecolor{lightgreen}{HTML}{90EE90}
\definecolor{lightred}{HTML}{FFB6C1}
\begin{figure}[h]
    \centering

    \subcaptionbox{Panoramic \rreplaced{input}{front} view at timestep 2}[1\columnwidth]{%
        \includegraphics[width=0.9\columnwidth, trim={0 900bp 0 0}, clip]{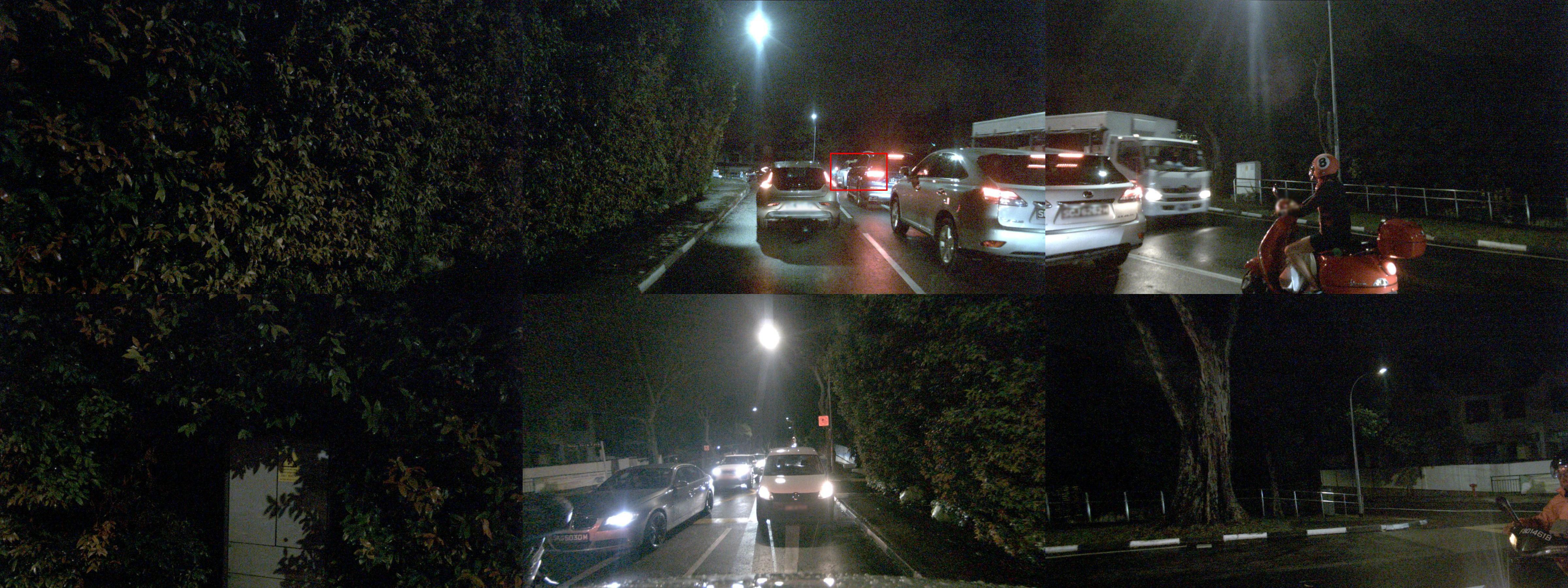}
    }
    \vspace{1pt} 


    \begin{minipage}[h]{0.50\columnwidth} 
        \centering
        \vspace{0pt}
        \subcaptionbox{BEV map}[1\linewidth]{%
            \includegraphics[width=\linewidth]{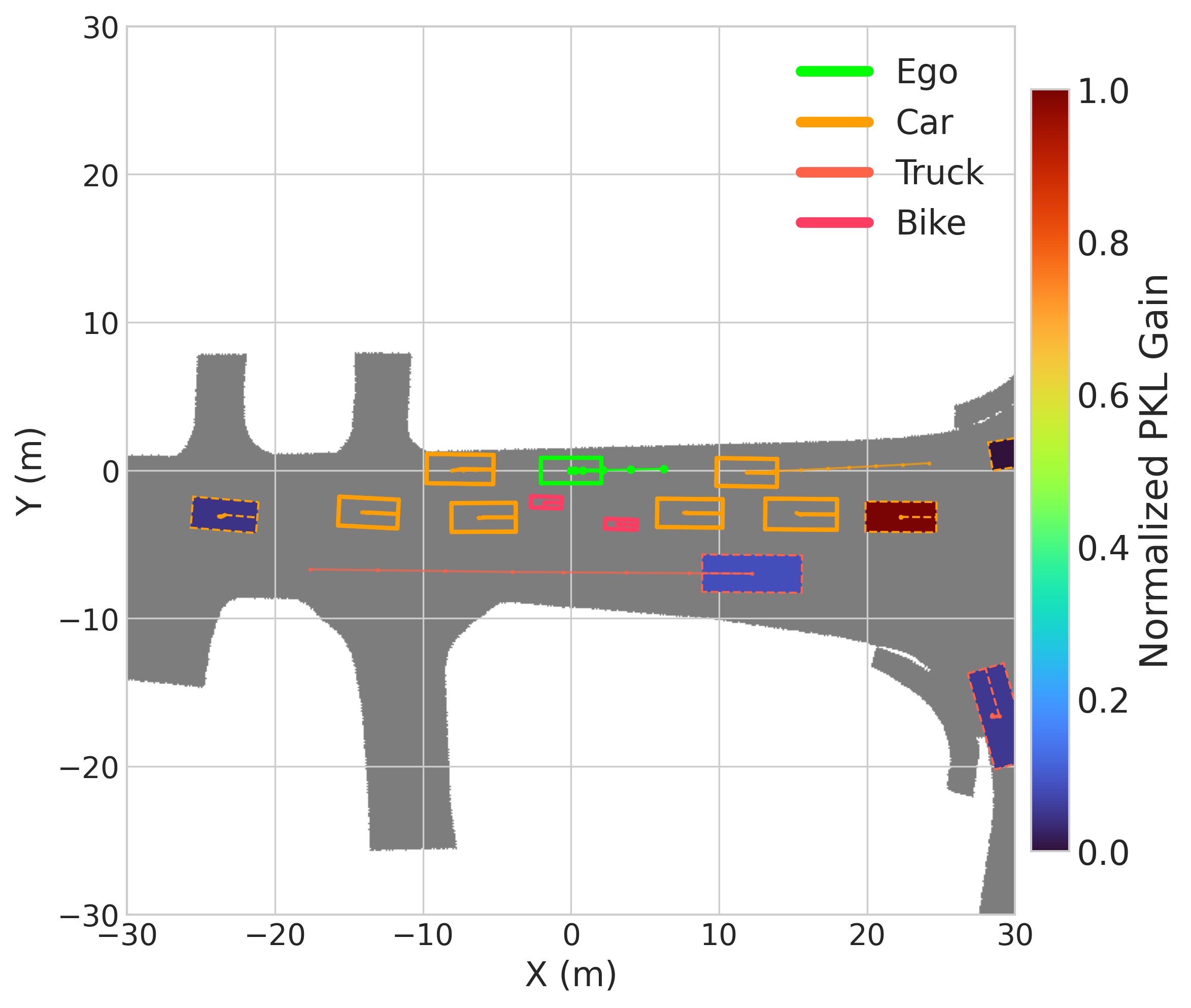}
        }
    \end{minipage}%
    \hfill 
\begin{minipage}[h]{0.5\columnwidth} 
    \centering
    \vspace{0pt}
    \subcaptionbox{Model predictions.}[1\linewidth]{%
        \tiny
        \setlength{\tabcolsep}{3pt} 
        \renewcommand{\arraystretch}{1.2}
        \begin{tabular}{@{} l c c c @{}}
            \toprule
             & & \multicolumn{2}{c}{\textbf{Random}} \\ 
            \cmidrule(lr){3-4} 
            \textbf{Field} & \textbf{PKL} & \textbf{Qwen} & \textbf{Intern} \\
            \midrule
            \texttt{class} & \cellcolor{lightgreen}car & \cellcolor{lightgreen}car & \cellcolor{lightgreen}car \\
            \texttt{action} & \cellcolor{lightgreen}stopped & \cellcolor{lightred}parked & \cellcolor{lightgreen}stopped \\
            \texttt{cam\_view} & \cellcolor{lightgreen}FRONT & \cellcolor{lightred}BACK & \cellcolor{lightred}BACK \\
            \texttt{vis\_cue} & \cellcolor{lightgreen}light\_sig & \cellcolor{lightred}partial & \cellcolor{lightred}partial \\
            \texttt{occ\_type} & \cellcolor{lightgreen}agent & \cellcolor{lightred}landmark & \cellcolor{lightgreen}agent \\
            \texttt{occ\_class} & \cellcolor{lightgreen}car & \cellcolor{lightred}vegetation & \cellcolor{lightgreen}car \\
            \texttt{priority} & \cellcolor{lightgreen}medium & \cellcolor{lightred}low & \cellcolor{lightgreen}medium \\
            \texttt{sector} & \cellcolor{lightgreen}Ahead & \cellcolor{lightred}B\_Left & \cellcolor{lightred}B\_Right \\
            \bottomrule
        \end{tabular}
    }
\end{minipage}

    \caption{A challenging nighttime scene. \rreplaced{The models (Qwen2.5-VL-7B and InternVL3.5-8B)}{Qwen-7B and Intern-8B}, when trained on randomly selected data, struggle with predicting priority and location compared to their PKL-guided counterparts. \rreplaced{Green and red highlights indicate correct and incorrect predictions, respectively.}{Green = correct, red = incorrect.}
    \vspace{-5pt}}
    \label{fig:abli-night}
\end{figure}

\subsubsection{Fine-Tuning Strategy Analysis}
Table~\ref{tab:sft-abliation} compares fine-tuning only the language model versus our full approach (LLM + alignment module). Improvements are modest but consistent for Qwen-7B, with gains across all metrics including the largest improvement in action score (+0.007). InternVL3.5-8B shows mixed results: accuracy and F1 improve marginally while precision slightly decreases, indicating that the benefit of aligner fine-tuning is model-dependent. Given the minimal computational overhead and consistent gains for Qwen, we recommend fine-tuning both components, though the LLM-only approach remains viable, particularly for architectures that show diminishing returns from aligner adaptation.
\begin{table}[h]
\centering
\caption{Fine-tuning comparison. Top row: LLM-only baseline. Bottom row: LLM+Aligner. Bold values indicate an improvement over the baseline (deltas in parentheses).}
\label{tab:sft-abliation}
\scriptsize
\setlength{\tabcolsep}{3pt}
\begin{tabular}{ll lll|l}
\toprule
 & & \multicolumn{3}{c}{\textbf{Average Class}} & \textbf{Action} \\
\cmidrule(lr){3-5} \cmidrule(lr){6-6}
\textbf{Model} & \textbf{Variant} & Acc. & Prec. & F1 & Avg. \\
\midrule
\multirow{2}{*}{Qwen-7B} & LLM & 0.538 & 0.528 & 0.508 & 0.428 \\
 & +Aligner & \textbf{0.543}{\scriptsize (+.005)} & \textbf{0.533}{\scriptsize (+.005)} & \textbf{0.514}{\scriptsize (+.006)} & \textbf{0.435}{\scriptsize (+.007)} \\
\midrule
\multirow{2}{*}{Intern-8B} & LLM & 0.564 & 0.497 & 0.494 & 0.440 \\
 & +Aligner & \textbf{0.565}{\scriptsize (+.001)} & 0.486{\scriptsize ($-$.011)} & \textbf{0.496}{\scriptsize (+.002)} & 0.440 \\
\bottomrule
\vspace{-10pt}
\end{tabular}
\end{table}

\subsubsection{Impact on Visible-Agent Performance}
\label{sec:visible-ablation}
To verify that fine-tuning on our occlusion-focused dataset does
not degrade performance on fully visible agents, we evaluate on the
MME-RealWorld~\cite{zhang2025mmerealworld} autonomous driving subset, which
contains predominantly unoccluded traffic scenarios. We compare base
models against their counterparts fine-tuned on our PKL-guided
dataset. As shown in Table~\ref{tab:mme-realworld}, DriveLM retains
identical reasoning while overall and perception drop by only
1-2\%. Qwen-3B shows even smaller degradation, with all scores
decreasing by less than 1\%. This confirms that our task-specific
fine-tuning introduces no meaningful catastrophic forgetting of
visible-agent capabilities across model families.\begin{table}[h]
\centering
\caption{Performance on the MME-RealWorld autonomous driving subset (deltas in parentheses).}
\label{tab:mme-realworld}
\footnotesize
\setlength{\tabcolsep}{6pt}
\begin{tabular}{l ccc}
\toprule
\textbf{Model} & \textbf{Overall} & \textbf{Reasoning} & \textbf{Perception} \\
\midrule
DriveLM & 0.482 & 0.403 & 0.511 \\
DriveLM(fine-tuned) & 0.474{\scriptsize ($-$.008)} & 0.403 & 0.500{\scriptsize ($-$.011)} \\
\midrule
Qwen-3B & 0.307 & 0.266 & 0.321\\
Qwen-3B(fine-tuned) & 0.305{\scriptsize ($-$.002)} & 0.265{\scriptsize ($-$.001)} & 0.320{\scriptsize ($-$.001)} \\
\bottomrule
\end{tabular}
\end{table}
\vspace{-10pt}

\subsection{Discussion}
Our results reveal several important insights:

\textbf{Task-Specific Training Dominates:} The large improvements from fine-tuning outweigh the benefits of model scale or domain adaptation, highlighting the importance of curated, task-specific data.\textbf{Planning-Aware Data Selection Matters:} PKL-guided selection provides $\sim$30\% better results than random selection, validating our approach of prioritizing occlusions that affect planning decisions.

\textbf{Architectural Efficiency:} Some model families (InternVL3.5) achieve comparable performance with fewer parameters, suggesting architectural innovations could further improve efficiency. Our fine-tuned models exhibit significant inference latency variations across architectures, with Qwen models achieving 47--79\,ms per sample, Gemma models requiring 74--99\,ms, and InternVL variants ranging from 86--323\,ms. These latencies suggest that smaller fine-tuned models are approaching real-time feasibility, particularly if the VLM runs as a parallel reasoning module at a lower frequency (e.g., 2--5\,Hz) than the main perception stack.\textbf{Action Understanding:} The improvements in action prediction indicate that models learn to infer hidden agent intentions from indirect visual evidence, that is a crucial capability for safe autonomous driving.\textbf{Downstream Utility:} The structured output format enables concrete integration pathways. The priority level and sector fields can directly trigger driver warnings in an ADAS system (e.g., ``High-priority hidden agent, Ahead-Left''). In multi-agent collaboration scenarios (V2V, drone-vehicle, V2X), the structured annotations provide a compact, standardized format for an ego vehicle to query another agent about planning-critical occlusions in its blind ~\cite{wu2025multiagentads}. Priority levels and sectors can also be converted to cost-map modifications for downstream planners.\subsection{Limitations and Future Work}
\label{sec:limitations}
Our investigation opens several important avenues for future work.
Driving datasets like nuScenes are often limited by uniform collection processes, which reduces their diversity and excludes many real-world edge cases. The resulting long-tail distribution makes safety-critical occlusion events rare. A promising future direction is both more targeted data collection to capture a wider variety of high-risk scenarios, and cross-dataset validation on benchmarks such as nuPlan and Waymo to establish external validity. Our PKL-guided framework is dataset-agnostic in principle, it requires only 3D annotations and a planner, making such transfer straightforward. Additionally, the inference latency of large VLMs remains a challenge for real-time deployment. While our results show the promise of smaller fine-tuned models (e.g., Qwen-3B at 47\,ms), further optimization is needed to meet strict latency requirements, such as running the VLM at a reduced frequency alongside the main perception stack or offloading to edge infrastructure.
Finally, while our manual audit and strict schema enforcement mitigate annotation errors, reliance on a single LLM may still imprint systematic biases onto the dataset. Future work could explore inter-model agreement using a diverse ensemble of annotators, or human-in-the-loop verification at scale, to further quantify and reduce label noise. Looking forward, a natural next step is closed-loop evaluation, where the structured annotations feed directly into a planner to measure safety improvements beyond static QA metrics. This work also enables novel applications, from multi-agent to ADAS systems.

\section{Conclusion}
\label{sec:conclusion}

In this work, we introduced a systematic framework to identify, annotate, and reason about planning-critical occluded agents in autonomous driving. By leveraging Planning KL-divergence (PKL) to curate a dataset of high-impact scenarios, we demonstrated that fine-tuning VLMs on this specialized data yields substantial performance gains across all model families and scales. Our results show that this task-specific, planning-aware training is more critical than model size, enabling smaller models to outperform much larger zero-shot counterparts and proving that a focus on important occlusions leads to more effective learning. This approach bridges a critical gap between perception and planning, paving the way for autonomous systems that can reason more intelligently about risks and navigate complex, partially-observable environments with greater safety and efficiency.

\bibliographystyle{unsrt}
\bibliography{refs}






\end{document}